\documentclass{article} % For LaTeX2e
\usepackage{iclr2027_conference,times}

\usepackage{amsmath,amsfonts,bm}

\def\eqref#1{equation~\ref{#1}}
\def\1{\bm{1}}

\DeclareMathAlphabet{\mathsfit}{\encodingdefault}{\sfdefault}{m}{sl}
\SetMathAlphabet{\mathsfit}{bold}{\encodingdefault}{\sfdefault}{bx}{n}

\usepackage[utf8]{inputenc} % allow utf-8 input
\usepackage[T1]{fontenc}    % use 8-bit T1 fonts
\usepackage{hyperref}       % hyperlinks
\usepackage{url}            % simple URL typesetting
\usepackage{booktabs}       % professional-quality tables
\usepackage{amsfonts}       % blackboard math symbols
\usepackage{amsmath}
\usepackage{nicefrac}       % compact symbols for 1/2, etc.
\usepackage{microtype}      % microtypography
\usepackage{xcolor}         % colors
\usepackage{multirow}
\usepackage{multicol}
\usepackage{graphicx}
\usepackage[table]{xcolor}
\usepackage{caption}
\usepackage{subcaption}
\usepackage{wrapfig}
\usepackage{bm}
\usepackage{mathtools}

\definecolor{mainblue}{RGB}{0, 75, 135}

\definecolor{Red}{rgb}{0.768, 0.054, 0.054}
\definecolor{Blue}{rgb}{0.43, 0.65, 0.86}
\definecolor{Green}{rgb}{0,0.4,0.7}
\definecolor{hotpink}{rgb}{1.0, 0.41, 0.71}
\definecolor{brown}{rgb}{0.59, 0.29, 0.0}
\definecolor{darkpastelgreen}{rgb}{0.01, 0.75, 0.24}
\definecolor{celestialblue}{rgb}{0.29, 0.59, 0.82}
\definecolor{ceruleanblue}{rgb}{0.16, 0.32, 0.75}
\definecolor{goldenrod}{rgb}{0.85, 0.65, 0.13}
\definecolor{navyblue}{rgb}{0.0, 0.0, 0.5}
\definecolor{coolgrey}{rgb}{0.55, 0.57, 0.67}
\definecolor{darkseagreen}{rgb}{0.56, 0.74, 0.56}
\definecolor{darkturquoise}{rgb}{0.0, 0.81, 0.82}
\definecolor{berryred}{rgb}{0.79, 0.25, 0.14}
\definecolor{teagreen}{rgb}{0.81,0.94,0.75}
\definecolor{lightgrey}{rgb}{0.5,0.5,0.5}
\definecolor{purple}{rgb}{0.35,0.25,0.55}
\hypersetup{
    colorlinks=true,
    citecolor=teal,
    linkcolor=Red,
    urlcolor=Green,
    }

\newcommand{\highlight}[1]{\cellcolor{mainblue!10}{\textbf{#1}}}
\newcommand{\up}[1]{{\tiny\color{darkseagreen}$\mathrlap{\uparrow #1}$}}
\newcommand{\down}[1]{{\tiny\color{berryred}$\mathrlap{\downarrow #1}$}}

\title{SLEM: Spectrum-Aware Latent Expert Merging}

\author{
    \textbf{Dohyeon Kim}\textsuperscript{1*}
    \quad
    \textbf{Bedionita Soro}\textsuperscript{1*}
    \quad
    \textbf{Bruno Andreis}\textsuperscript{3,4}
    \quad
    \textbf{Sung Ju Hwang}\textsuperscript{1,2}
    \\[2pt]
    \textsuperscript{1}KAIST
    \quad
    \textsuperscript{2}DeepAuto.ai
    \quad
    \textsuperscript{3}University of Oxford
    \quad
    \textsuperscript{4}Slater Labs
    \\[2pt]
    {\fontsize{8.7pt}{11pt}\selectfont \texttt{\{dhkim011030, sorobedio, sjhwang82\}@kaist.ac.kr},\quad\texttt{bruno.andreis@eng.ox.ac.uk}}
}

\iclrfinalcopy % Uncomment for camera-ready version, but NOT for submission.
\begin{document}

\maketitle
\lhead{Preprint.}

\begin{abstract}
Mixture-of-Experts (MoE) models scale model capacity efficiently by activating only a small subset of experts for each input, while still requiring all parameters to be stored in memory, resulting in substantial memory overhead. Expert merging techniques have recently emerged as a promising compression paradigm that reduces this overhead by consolidating multiple experts into a smaller set. However, existing approaches have two key limitations: (i) direct convex interpolation in parameter space is not generally guaranteed to yield meaningful interpolation of expert functions; and (ii) estimating expert similarity or merging coefficients often relies on calibration data, introducing dependence on the calibration distribution. To address these limitations, we propose \textbf{Spectrum-Aware Latent Expert Merging (SLEM)}, a calibration-free expert merging framework that occurs in a structured latent representation of expert weights using a VAE. SLEM groups experts within and across layers based on latent similarities and derives merging coefficients from expert weight spectra. It then performs geometry-aware latent interpolation and decodes the resulting representations into merged expert weights. Compared with existing expert merging methods, SLEM achieves comparable or better performance on general knowledge benchmarks and higher average performance on generation benchmarks involving mathematical reasoning and code generation.

\end{abstract}

% \blfootnote{*Equal contribution.}

\section{Introduction} \label{sec:introduction}

Mixture-of-Experts (MoE) models scale capacity efficiently by activating only a small subset of experts for each token~\citep{shazeer2017, survey}. This sparsity reduces per-token computation, but not memory usage: all expert parameters must remain stored regardless of how frequently they are activated. As a result, the memory footprint scales with the total number of experts rather than the number used for each token. Expert pruning and expert merging address this mismatch by either removing experts or consolidating multiple experts into a smaller set~\citep{merginginone, mergecompressdemystifyefficient, miao2025mergemoeefficientcompressionmoe}.

Expert merging is particularly attractive because it seeks to retain information from the experts it consolidates rather than simply discarding them. Existing approaches, however, face two limitations. First, experts are typically combined directly in parameter space, although linear interpolation between neural-network parameters is not generally guaranteed to correspond to meaningful interpolation between the functions they represent. Neuron alignment~\citep{mergecompressdemystifyefficient}, curvature-aware merging~\citep{nguyen2025camex}, and subspace decomposition~\citep{li2026submoe} mitigate this issue, but the resulting merge remains tied to the original parameter space. Second, deciding which experts to merge and how strongly they should contribute often relies on calibration data through routing frequencies, routing patterns, activations, or expert outputs~\citep{mergecompressdemystifyefficient, HC-SMoE, li2026submoe}. Such statistics make the resulting compression decisions dependent on the calibration distribution. Improving the merging rule alone may therefore be insufficient: both the representation in which experts are combined and the information used to guide compression need to be reconsidered.

\begin{figure}[t]
    \includegraphics[width=\linewidth]{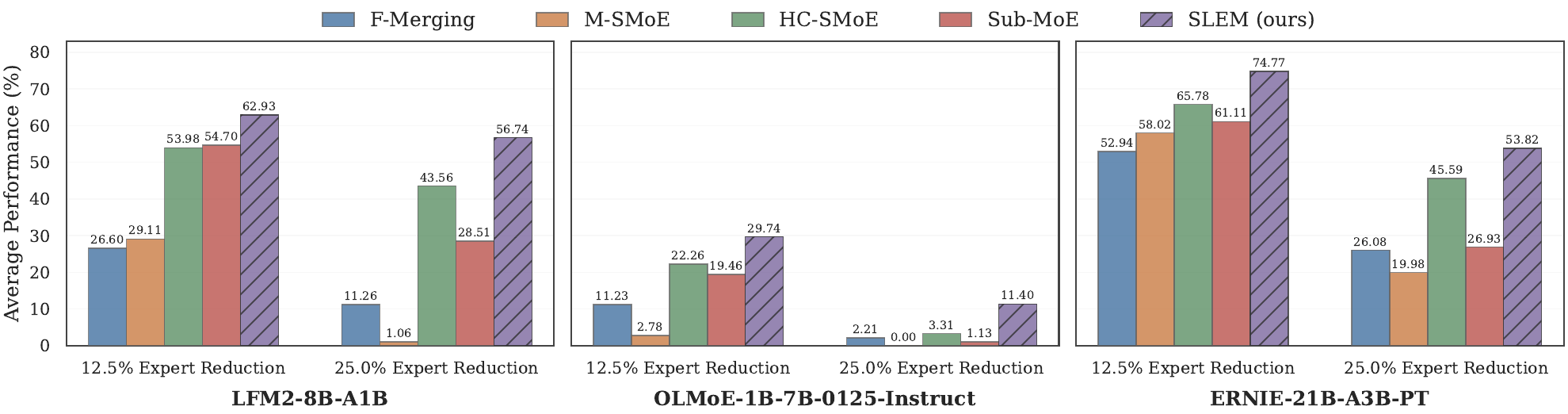}
    \caption{Average performance comparison of expert merging methods on generation-based benchmarks involving mathematical reasoning and code generation {\small(GSM8K, MATH500, MBPP, and HumanEval)}. SLEM consistently achieves the highest average performance under all evaluated settings.}
    \label{fig:generation_performance}
\end{figure}
These limitations motivate reformulating expert merging as a weight-space learning problem~\citep{weightspace}, shifting the merging domain from raw parameter space to a learned latent weight space~\citep{soro2026lsmerge} in which MoE experts may admit a geometry better suited to merging without calibration data. Building on this view, we propose \textbf{Spectrum-Aware Latent Expert Merging (SLEM)}, a calibration-free framework that merges experts in a structured latent representation learned with a variational autoencoder (VAE)~\citep{higgins2017betavae}. SLEM encodes expert weights, groups experts by latent similarity, aligns subnetworks in the same latent space, performs geometry-aware interpolation, and decodes the result into a merged expert, with subnetwork-specific merging coefficients derived from the spectral structure of the pretrained weights. It therefore requires neither calibration data nor additional forward passes: latent geometry replaces direct parameter-space interpolation, while spectral structure replaces input-dependent routing statistics with quantities computed directly from pretrained expert weights.

Our contributions are summarized as follows:
\begin{enumerate}
\item We propose \textbf{Spectrum-Aware Latent Expert Merging (SLEM)}, which introduces an Expert VAE to learn latent representations of expert weights and performs geometry-aware expert interpolation in the learned latent space.

\item  We introduce an expert merging strategy combining spectrum-aware merging coefficients with latent-similarity-based expert grouping, enabling fully calibration-free expert merging without additional data.

\item We extend expert merging beyond within-layer constraints and enable cross-layer expert merging by representing experts from all MoE layers in a shared latent space.

\item We demonstrate that SLEM achieves the highest average performance on mathematical reasoning and code generation benchmarks, with ablations validating latent-space merging, cross-layer merging, and spectrum-aware coefficients.

\end{enumerate}

Taken together, SLEM provides a new perspective on expert merging by changing both interpolation and merging decisions from input-dependent statistics toward the intrinsic structure of expert weights. 

\section{Related Work}

\paragraph{MoE Compression and Expert Merging.}
One line of work reduces the memory and inference cost of MoE models by pruning redundant parameters or experts~\citep{survey,moes-survey}, scoring them with weight magnitude, activations and router weights~\citep{MoE-Pruner}, approximate Shapley values~\citep{huang2025discovering}, learned per-layer ratios~\citep{DiEP}, cross-domain behavior clusters~\citep{Mosaic}, or informative tokens~\citep{zong2026less}, replacing redundant experts with lightweight novices~\citep{MoNE}, or exploiting finer-grained parameter structure through low-rank matrix sharing~\citep{svd-moe}, a shared base with low-rank deltas~\citep{delta}, and cross-matrix micro-experts~\citep{CAMERA}.
Expert merging is complementary, since it preserves information from the experts it consolidates rather than discarding them: MEO approximates several activated experts with one~\citep{merginginone}, M-SMoE aligns expert neurons and derives groups and coefficients from routing statistics~\citep{mergecompressdemystifyefficient}, HC-SMoE clusters experts by output similarity~\citep{HC-SMoE}, Sub-MoE extracts shared subspaces via joint SVD~\citep{li2026submoe}, NAMEx obtains merging weights from a Nash bargaining problem~\citep{ICLR2026_ef9f16cd}, and PuzzleMoE separates shared from expert-specific parameters for sparse element-wise merging~\citep{zhao2026puzzlemoe}.
In all of these, the merged expert is still constructed from the original parameters or their explicit decompositions, and grouping and weighting rely on routing frequencies, expert outputs, activations, or calibration examples.

\begin{figure}[t]
    \includegraphics[width=\linewidth]{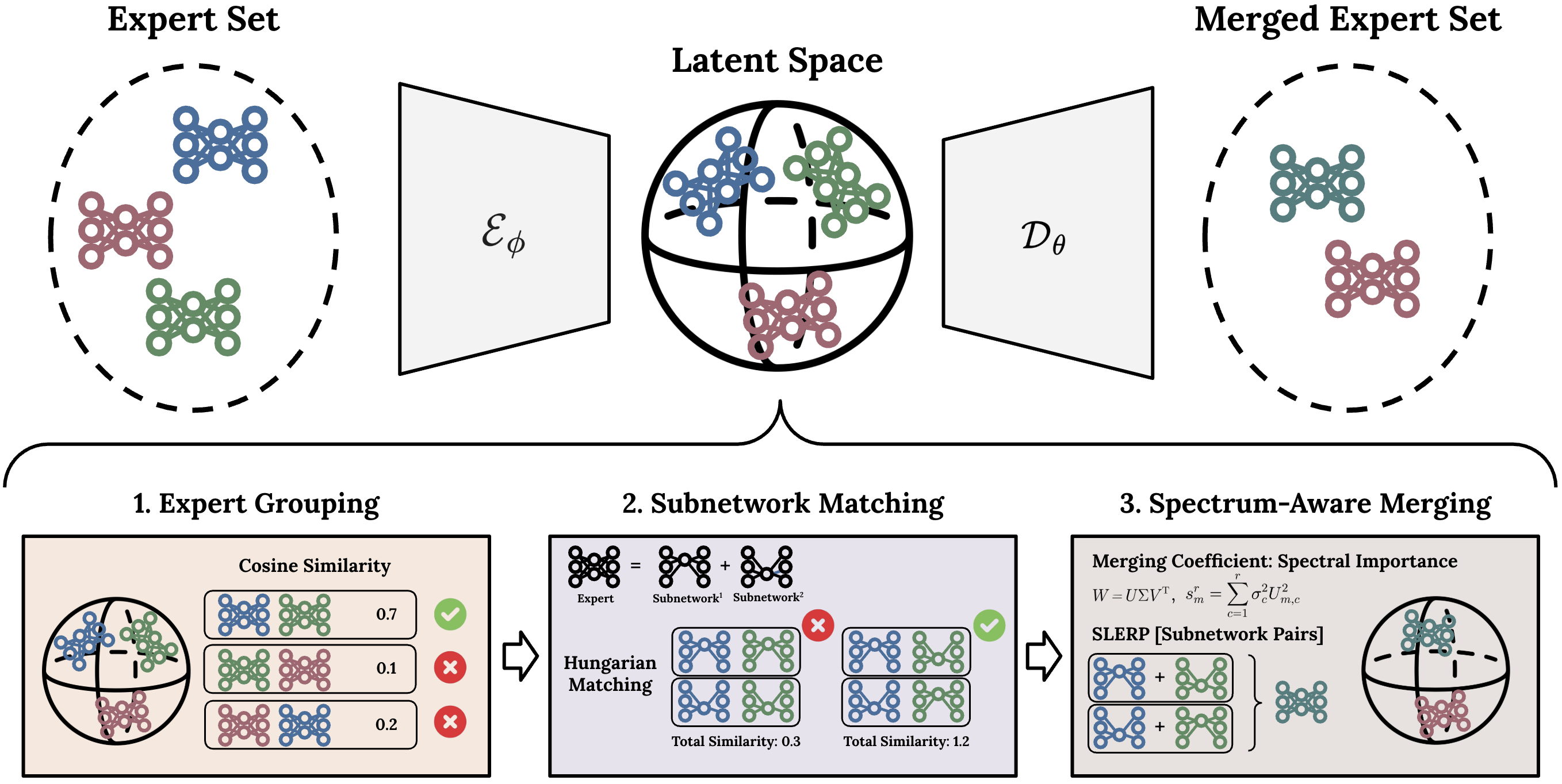}
    \caption{Overall framework of Spectrum-Aware Latent Expert Merging (SLEM).}
    \label{fig:SLEM_framework}
\end{figure}

\paragraph{Weight-Space Learning.}
A growing line of work treats trained network parameters as a data modality in their own right~\citep{weightspace,wang2026position}.
Early efforts learn representations over populations of trained models to predict model characteristics~\citep{NEURIPS2021_89562dcc}. Later work extends such representations to recurrent weight matrices~\citep{herrmann2024learningusefulrepresentationsrecurrent} and to downstream domains such as geospatial learning, where representations are learned from trained models rather than raw data~\citep{geosane}.
Because network functions are invariant to neuron permutations, a parallel line designs permutation-equivariant architectures that process weights directly~\citep{zhou2023permutationequivariantneuralfunctionals,NEURIPS2023_f4757db8}, recently extended to the symmetries of Transformer weights~\citep{ICLR2025_29962c2c}.
On the generative side, hyper-representations embed weights from model populations and generate new parameter configurations from them~\citep{hyper-representations}, followed by diffusion-based generators~\citep{soro2025diffusionbased}, flow matching with re-basin alignment to handle permutation symmetry~\citep{deepweightflow}, and width-agnostic generation over structurally aligned weight spaces~\citep{nnit}.
Weight-space operations also underlie model editing and merging. Task arithmetic combines fine-tuning deltas directly in parameter space~\citep{ilharco2023editing}, whereas LS-Merge, closest to our setting, encodes pretrained language-model weights with a Transformer-based variational autoencoder, aligns heterogeneous models in the learned representation, and interpolates them in latent space before decoding back to parameters~\citep{soro2026lsmerge}.
While these methods establish weight space as a useful domain for representation, generation, and model-level manipulation, we study learned weight representations at the granularity of individual MoE experts and use their geometry explicitly for compression.
Rather than merging complete models or interpolating experts directly in parameter space, we learn latent representations of individual experts and combine their latent geometry with spectral structure to determine expert groupings and merging coefficients without calibration data.

\begin{figure}[t]
    \centering

    % ---------- Row 1 ----------
    \begin{subfigure}{0.32\linewidth}
        \centering
        \includegraphics[width=\linewidth]{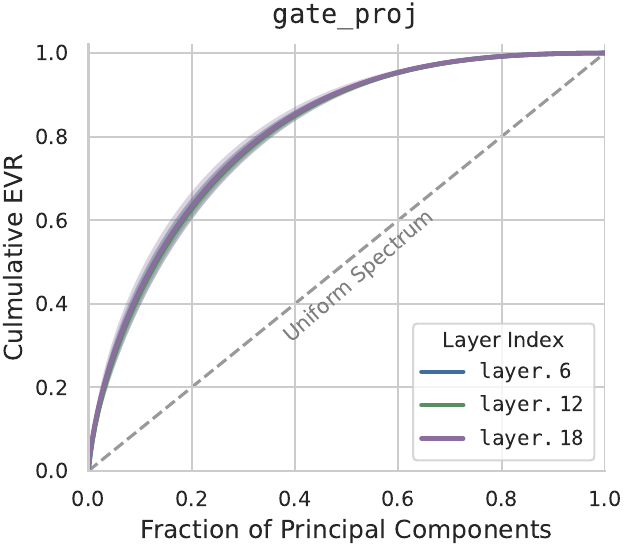}
    \end{subfigure}
    \hfill
    \begin{subfigure}{0.32\linewidth}
        \centering
        \includegraphics[width=\linewidth]{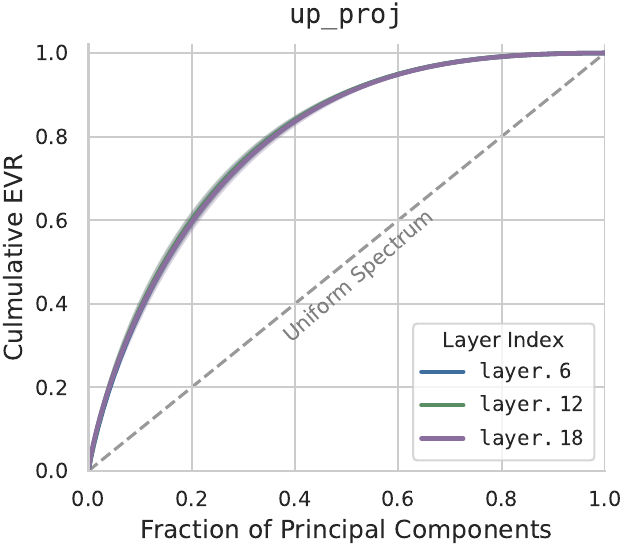}
    \end{subfigure}
    \hfill
    \begin{subfigure}{0.32\linewidth}
        \centering
        \includegraphics[width=\linewidth]{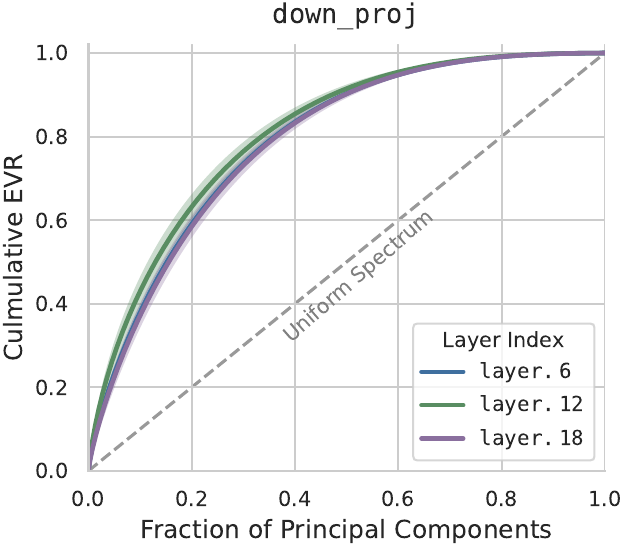}
    \end{subfigure}

    \caption{Culmulative explained variance ratio (EVR) of expert projection weights across layers in LFM2-8B-A1B. The EVR curves show that variance is concentrated along a subset of principal directions across all projection types, motivating the use of dominant spectral structure for estimating subnetwork importance. Results for the other evaluated models are provided in Appendix~\ref{app:culmulative_evr}.}
    \label{fig:expert_evr_curve}
\end{figure}

\section{Spectrum-Aware Latent Expert Merging (SLEM)} \label{sec:method}
Consider a pretrained Mixture-of-Experts (MoE) model with $L$ MoE layers, where layer $\ell$ contains $K$ experts $\{E_1^{(\ell)},\ldots,E_K^{(\ell)}\}$, of which only $k\ll K$ are activated for each token.
Unlike conventional within-layer expert merging, SLEM allows expert pairs to be selected both within and across MoE layers.
Accordingly, our objective is to reduce the expert-memory footprint without retraining by replacing an expert pair $\mathcal{G}=\{E_i^{(\ell)},E_j^{(\ell')}\}$, where $\ell$ and $\ell'$ may be identical or different, with a single merged expert.
Each pairwise merge therefore reduces the number of stored expert parameter sets by one.
Rather than combining the corresponding expert weights directly in weight space, we treat them as structured data, map them to learned latent representations, merge in that space, and decode the result back into a single expert. For notational simplicity, we omit layer indices throughout, while allowing $E_i$ and $E_j$ to originate from either the same or different MoE layers.

\subsection{Hyperspherical Expert VAE} \label{sec:vae_training}
Following LS-Merge~\citep{soro2026lsmerge}, we learn latent representations of expert weights with transformer-based variational autoencoders (VAEs). 
To reflect the internal structure of MoE experts, we design an Expert VAE that jointly models the three projection types
$p\in{\small\mathcal{P}=\{\texttt{gate},\texttt{up},\texttt{down}\}}$. The training data consist solely of the non-shared expert weights of the pretrained model, obtained by splitting the fused {\small\texttt{gate\_up\_proj}} weights of each expert $E_i$ into gate and up projections and retaining the corresponding {\small\texttt{down\_proj}} weights. 
Each projection weight is represented as a sequence along the hidden-unit dimension, since a gated feed-forward expert with intermediate width $D_{\mathrm{MoE}}$ decomposes into contributions from its hidden units, which we refer to as subnetworks,
\begin{equation}
E_i(x)
=
\sum_{m=1}^{D_{\mathrm{MoE}}}E_{i,m}(x),
\qquad
E_{i,m}(x)
=
\varphi\!\left(g_{i,m}^{\top}x\right)
\left(u_{i,m}^{\top}x\right)d_{i,m},
\label{eq:subnet}
\end{equation}
where $g_{i,m}$ and $u_{i,m}$ are the $m$-th rows of the gate and up projections respectively, $d_{i,m}$ is the $m$-th column of the down projection, and $\varphi$ is the activation function.
The triplet $(g_{i,m},u_{i,m},d_{i,m})$ defines a subnetwork index common to the three projections.
Organizing the projections by this index gives $W_{i,\texttt{gate}}=G_i$, $W_{i,\texttt{up}}=U_i$, and $W_{i,\texttt{down}}=D_i^{\top}$, all in $\mathbb{R}^{D_{\mathrm{MoE}}\times D}$, where the $m$-th row of $W_{i,p}$ holds the component of subnetwork $m$ in projection $p\in\mathcal{P}$ and $D$ is the model hidden dimension.
A single training sample for each expert is constructed by concatenating the three projection weights along the sequence dimension,
$W_i=\left[W_{i,\texttt{gate}};W_{i,\texttt{up}};W_{i,\texttt{down}}\right]
\in \mathbb{R}^{3D_{\mathrm{MoE}}\times D}$.
We use type-specific projections to distinguish the gate, up, and down segments while jointly modeling their relationships within each expert. In addition, we omit positional encodings because the absolute ordering of subnetworks has no canonical functional meaning, as reflected in \eqref{eq:subnet}.

\paragraph{Hyperspherical prior and objective.}
The prior toward which the posterior is regularized determines the geometry in which expert representations are later compared and interpolated, so its choice is central to our setting.
\citet{hyper-representations} observed that learned weight embeddings tend to occupy a shell in high-dimensional space, which is poorly matched by the Gaussian prior commonly used in VAEs.
We therefore adopt a hyperspherical VAE~\citep{davidson2018hyperspherical} with von Mises--Fisher (vMF) prior and use a fixed concentration $\kappa$ following prior work~\citep{xu-durrett-2018-spherical}. For an expert weight $W$, the encoder defines $q_{\phi}(\mathbf{z}\mid W) = \mathrm{vMF}\!\left( \mu_{\phi}(W),\kappa \right)$, while the prior is parameterized as
$p_{\psi}(\mathbf{z}) = \mathrm{vMF}\!\left( \nu,\kappa \right)$,
where $\nu$ is a learnable prior direction. We optimize
\begin{equation}
\mathcal{L}
=
\min_{\theta,\phi,\psi}
\;
\mathbb{E}_{\mathbf{z}\sim q_{\phi}(\mathbf{z}\mid W)}
\left[
\left\|
W-\mathcal{D}_{\theta}(\mathbf{z})
\right\|_{F}^{2}
\right]
+
\beta
D_{\mathrm{KL}}
\left(
q_{\phi}(\mathbf{z}\mid W)
\,\middle\|\,
p_{\psi}(\mathbf{z})
\right),
\label{eq:expert_elbo}
\end{equation}
subject to
$\|\mathbf{z}\|_{2}
=
\|\mu_{\phi}(W)\|_{2}
=
\|\nu_{\ell}\|_{2}
=1$,
with $\beta$ controlling the trade-off between reconstruction fidelity and latent regularity~\citep{higgins2017betavae}.
Both the sampled latent and the vMF mean directions therefore lie on $\mathbb{S}^{d-1}$; we denote the latent representation of subnetwork $m$ of expert $i$ in projection $p$ as $\mathbf{z}_{i,p}^{m}$.

\subsection{Latent Expert Merging} \label{sec:SLEM}

Given the learned latent space, SLEM (i) identifies expert pairs and aligns their subnetworks using latent similarity, (ii) assigns subnetwork-specific merging coefficients based on spectral importance, and (iii) performs geometry-aware latent interpolation, followed by decoding the merged expert.

\paragraph{Latent subnetwork similarity.}
Since the latent representations lie on the unit hypersphere, their geometry is characterized by direction, which makes cosine similarity a natural choice. We define the compatibility between subnetwork $m$ of expert $i$ and subnetwork $n$ of expert $j$ as
\begin{equation}
C_{m,n}^{(i,j)}
=
\frac{1}{|\mathcal{P}|}
\sum_{p\in\mathcal{P}}
\frac{
(\mathbf{z}_{i,p}^{m})^{\top}\mathbf{z}_{j,p}^{n}
}{
\|\mathbf{z}_{i,p}^{m}\|_{2}
\|\mathbf{z}_{j,p}^{n}\|_{2}
}.
\label{eq:latent-sim}
\end{equation}
Based on \eqref{eq:latent-sim}, we define the similarity of the two experts $i,j$ as the mean compatibility over all subnetwork pairs, $\mathcal{S}(i,j) = \frac{1}{D_{\mathrm{MoE}}^{2}}\sum_{m=1}^{D_{\mathrm{MoE}}}\sum_{n=1}^{D_{\mathrm{MoE}}} C_{m,n}^{(i,j)}$.
Expert pairs with higher \(\mathcal{S}(i,j)\) are considered more compatible and are therefore prioritized for grouping. We form pairwise merge groups by greedy one-to-one matching on $\mathcal{S}$: at each step we take the most similar remaining pair, assign it to a merge group, and remove both experts from consideration, until the desired number of pairs is reached.
We additionally impose a layer-wise limit on the number of merged experts to prevent merge groups from being overly concentrated in a small subset of layers.

Within a group, subnetworks still have no canonical ordering across experts and must be aligned before interpolation. Hungarian matching of hidden units substantially improves expert merging~\citep{mergecompressdemystifyefficient}, so we follow this approach but define the criterion in latent space, seeking the permutation $\pi$ of $[D_{\mathrm{MoE}}]$ that maximizes total compatibility defined as $\pi^{\star}=\arg\max_{\pi}\sum_{m=1}^{D_{\mathrm{MoE}}}C_{m,\pi(m)}^{(i,j)}.$
The resulting permutation is applied consistently across the three projections, preserving the shared subnetwork index of \eqref{eq:subnet}.

\paragraph{Spectral importance as merging coefficients.}
As shown in Figure~\ref{fig:expert_evr_curve}, subnetworks do not contribute equally to the dominant structure of an expert, so we set their coefficients from the spectrum of the pretrained weights $W_{i,p}=A_{i,p}\Sigma_{i,p}B_{i,p}^{\top}$, where $\sigma_{i,p,c}$ is the $c$-th singular value and $A_{i,p,mc}$ the $(m,c)$ entry of $A_{i,p}$.
We quantify the spectral importance of subnetwork $m$ as
\begin{equation}
\tilde{s}_{i,p,m}^{\,r}
=
\sum_{c=1}^{r}
\sigma_{i,p,c}^{2}A_{i,p,mc}^{2},
\quad
s_{i,p,m}^{\,r}
=
\frac{
\tilde{s}_{i,p,m}^{\,r}
}{
\sum_{n}
\tilde{s}_{i,p,n}^{\,r}
},
\quad
s_{i,m}^{\,r}
=
\frac{1}{|\mathcal{P}|}
\sum_{p\in\mathcal{P}}s_{i,p,m}^{\,r},
\label{eq:importance}
\end{equation}
which measures the amount of spectral energy in the leading $r$ singular directions attributable to subnetwork $m$. A larger value therefore indicates that the subnetwork contributes more strongly to the dominant structure of the expert. For a matched pair $(m,\pi^{\star}(m))$, we then determine the merging coefficient from their relative spectral importance, assigning a larger contribution to the more spectrally important subnetwork:
\begin{equation}
\alpha_m
=
\frac{
s_{i,m}^{\,r}
}{
s_{i,m}^{\,r}+s_{j,\pi^{\star}(m)}^{\,r},
},
\quad
\alpha_{\pi^{\star}(m)}
=
\frac{
s_{j,\pi^{\star}(m)}^{\,r}
}{
s_{i,m}^{\,r}+s_{j,\pi^{\star}(m)}^{\,r},
}
=
1 - \alpha_m.
\label{eq:coefficients}
\end{equation}
These coefficients replace uniform averaging and routing-frequency weighting, and the rank $r$, which controls how much of the leading spectrum is retained, is the only free parameter of the rule.

\paragraph{Latent expert interpolation.}
Because the representations are directional, we merge each aligned pair with spherical linear interpolation as $\widetilde{\mathbf{z}}_{ij}^{\,m}=\operatorname{SLERP}(\mathbf{z}_{i}^{m},\mathbf{z}_{j}^{\pi^{\star}(m)};1-\alpha_m)$, where $\operatorname{SLERP}(a,b;t)$ interpolates from $a$ at $t=0$ to $b$ at $t=1$, so a larger $\alpha_m$ weights the subnetwork from expert $i$ more heavily.
The interpolated latents are assembled into $\widetilde{Z}_{ij}$ and decoded as $\widetilde{W}_{ij}=\mathcal{D}_{\theta}(\widetilde{Z}_{ij})$.
The resulting expert then replaces the pair in the compressed MoE layer.
\begin{table}[t]
\caption{Downstream performance comparison with existing expert merging methods. All merging methods are evaluated under the same expert reduction rate. The best and second-best results within each expert reduction setting are highlighted in \textbf{bold} and \underline{underlined}, respectively.}
\centering
\resizebox{\linewidth}{!}{
\begin{tabular}{c|l c c c c c c c c c}

\toprule
\multicolumn{1}{c}{} & & \multicolumn{4}{c}{\small General Knowledge (MCQA)} & \multicolumn{5}{c}{\small Mathematics \& Coding (Generation)} \\
\cmidrule(lr){3-6} \cmidrule(lr){7-11}
\multicolumn{1}{c}{\textbf{Expert Reduction}} & \textbf{Method} &  \;\textbf{MMLU}\; & \;\textbf{ARC-E}\; & \;\textbf{ARC-C}\; & \textbf{Average $\uparrow$} & \textbf{GSM8K} & \textbf{MATH500} &\textbf{MBPP} & \textbf{HumanEval} & \textbf{Average $\uparrow$} \\

%%%%%%%%%%%%%%%%%%%%%%%%%%%%
%%% LFM2-8B-A1B
%%%%%%%%%%%%%%%%%%%%%%%%%%%%
\midrule
\midrule
\multicolumn{11}{c}{\small LFM2-8B-A1B} \\

\midrule
\multirow{2}{*}{\small 0.0\%} & Vanilla & \small$63.11$ & \small$75.59$ & \small$47.63$ & \small$62.11$ & \small$81.73$ & \small$72.40$ & \small$45.60$ & \small$70.73$ & \small$67.62$ \\
& VAE-reconstructed & \small$62.93_{\mathrlap{\pm0.21}}$ & \small$75.61_{\mathrlap{\pm0.45}}$ & \small$47.41_{\mathrlap{\pm0.61}}$ & \small$61.98_{\mathrlap{\pm0.32}}$ & \small$82.35_{\mathrlap{\pm0.30}}$ & \small$72.20_{\mathrlap{\pm0.64}}$ & \small$45.52_{\mathrlap{\pm1.14}}$ & \small$71.83_{\mathrlap{\pm3.17}}$ & \small$67.97_{\mathrlap{\pm0.72}}$ \\
\midrule
\multirow{5}{*}{\small 12.5\%} & F-Merging & \small$55.64$ & \small$68.18$ & \small$42.32$ & \small$55.38$ & \small$69.14$ & \small$30.60$ & \small$03.00$ & \small$03.66$ & \small$26.60$ \\
& M-SMoE & \small$54.72$ & \underline{\small$72.97$} & \underline{\small$44.11$} & \underline{\small$57.27$} & \small$57.24$ & \small$20.40$ & \small$13.20$ & \small$25.61$ & \small$29.11$ \\
& HC-SMoE & \underline{\small$57.45$} & \small$69.82$ & \small$43.87$ & \small$57.05$ & \small$64.97$ & \small$56.00$ & \underline{\small$37.00$} & \underline{\small$57.93$} & \small$53.98$ \\
& Sub-MoE & \small$56.28$ & \small$69.60$ & \small$42.27$ & \small$56.05$ & \underline{\small$70.13$} & \underline{\small$59.60$} & \small$34.80$ & \small$54.27$ & \underline{\small$54.70$} \\

\cline{2-11}
& \rule{0pt}{2.3ex}\highlight{SLEM (\bm{$r$}=14)} & \highlight{\small\textbf{58.37}} & \highlight{\small\textbf{74.79}} & \highlight{\small\textbf{47.87}} & \highlight{\small\textbf{60.34}} & \highlight{\small\textbf{77.10}} & \highlight{\small\textbf{67.60}} & \highlight{\small\textbf{44.20}} & \highlight{\small\textbf{62.80}} & \highlight{\small\textbf{62.93}} \\

\midrule
\multirow{6}{*}{\small 25.0\%} & F-Merging & \small$47.72$ & \small$65.03$ & \small$38.23$ & \small$50.32$ & \small$37.83$ & \small$06.40$ & \small$00.20$ & \small$00.61$ & \small$11.26$ \\
& M-SMoE & \small$46.07$ & \underline{\small$66.89$} & \underline{\small$40.31$} & \underline{\small$51.09$} & \small$02.05$ & \small$02.20$ & \small$00.00$ & \small$00.00$ & \small$01.06$ \\
& HC-SMoE & \underline{\small$50.98$} & \small$60.47$ & \small$37.15$ & \small$49.53$ & \underline{\small$54.13$} & \underline{\small$45.20$} & \underline{\small$30.40$} & \underline{\small$44.51$} & \underline{\small$43.56$} \\
& Sub-MoE & \small$49.91$ & \small$64.82$ & \small$37.75$ & \small$50.83$ & \small$43.59$ & \small$34.20$ & \small$08.20$ & \small$28.05$ & \small$28.51$ \\

\cline{2-11}
& \rule{0pt}{2.3ex}\highlight{SLEM (\bm{$r$}=14)} & \highlight{\small$51.10$} & \highlight{\small\textbf{70.62}} & \highlight{\small\textbf{42.75}} & \highlight{\small\textbf{54.82}} & \highlight{\small\textbf{70.36}} & \highlight{\small$58.80$} & \highlight{\small\textbf{36.20}} & \highlight{\small\textbf{61.59}} & \highlight{\small\textbf{56.74}} \\
\rowcolor{mainblue!10}\cellcolor{white} & $\;\llcorner$ within-layer merging & \small\textbf{51.29} & \small67.89 & \small41.47 & \small53.55 & \small68.16 & \small\textbf{59.00} & \small31.40 & \small59.76 & \small54.58 \\

%%%%%%%%%%%%%%%%%%%%%%%%%%%%%%%%%%
%%% OLMoE-1B-7B-0125-Instruct
%%%%%%%%%%%%%%%%%%%%%%%%%%%%%%%%%%
\midrule
\midrule
\multicolumn{11}{c}{\small OLMoE-1B-7B-0125-Instruct} \\

\midrule
\multirow{2}{*}{\small 0.0\%} & Vanilla & \small$53.45$ & \small$76.48$ & \small$47.17$ & \small$59.03$ & \small$68.99$ & \small$23.00$ & \small$29.60$ & \small$35.37$ & \small$39.24$ \\
& VAE-reconstructed & \small$53.21_{\mathrlap{\pm0.10}}$ & \small$76.36_{\mathrlap{\pm0.56}}$ & \small$47.39_{\mathrlap{\pm0.75}}$ & \small$58.98_{\mathrlap{\pm0.32}}$ & \small$68.89_{\mathrlap{\pm0.89}}$ & \small$23.64_{\mathrlap{\pm1.08}}$ & \small$28.40_{\mathrlap{\pm0.89}}$ & \small$33.54_{\mathrlap{\pm1.02}}$ & \small$38.62_{\mathrlap{\pm0.42}}$ \\

\midrule
\multirow{5}{*}{\small 12.5\%} & F-Merging & \small$45.13$ & \small$66.12$ & \small$38.91$ & \small$50.05$ & \small$41.32$ & \small$03.60$ & \small$00.00$ & \small$00.00$ & \small$11.23$ \\
& M-SMoE & \small$43.56$ & \small$63.43$ & \small$35.75$ & \small$47.58$ & \small$07.51$ & \small$03.60$ & \small$00.00$ & \small$00.00$ & \small$02.78$ \\
& HC-SMoE & \small\textbf{50.58} & \underline{\small$75.49$} & \small\textbf{48.02} & \small\textbf{58.03} & \underline{\small$56.03$} & \underline{\small$09.40$} & \underline{\small$12.00$} & \underline{\small$11.59$} & \underline{\small$22.26$} \\
& Sub-MoE & \small$46.26$ & \small$69.82$ & \small$39.64$ & \small$51.91$ & \small$48.45$ & \small$07.20$ & \small$10.60$ & \underline{\small$11.59$} & \small$19.46$ \\

\cline{2-11}
& \rule{0pt}{2.3ex}\highlight{SLEM (\bm{$r$}=16)} & \highlight{\underline{\small$47.37$}} & \highlight{\small\textbf{75.80}} & \highlight{\underline{\small$45.90$}} & \highlight{\underline{\small$56.36$}} & \highlight{\small59.74} & \highlight{\small14.20} & \highlight{\small20.00} & \highlight{\small25.00} & \highlight{\small29.74} \\

\midrule
\multirow{5}{*}{\small 25.0\%} & F-Merging & \small$32.69$ & \small$62.71$ & \small$35.32$ & \small$43.58$ & \small$05.23$ & \small$03.60$ & \small$00.00$ & \small$00.00$ & \small$02.21$ \\
& M-SMoE & \small$23.16$ & \small$42.10$ & \small$26.33$ & \small$30.53$ & \small$00.00$ & \small$00.00$ & \small$00.00$ & \small$00.00$ & \small$00.00$ \\
& HC-SMoE & \small\textbf{43.52} & \underline{\small$69.41$} & \small\textbf{41.67} & \small\textbf{51.54} & \underline{\small$06.44$} & \underline{\small$03.80$} & \underline{\small$02.40$} & \underline{\small$00.61$} & \underline{\small$03.31$} \\
& Sub-MoE & \underline{\small$33.56$} & \small$39.93$ & \small$22.87$ & \small$32.12$ & \small$00.91$ & \small$01.60$ & \small$02.00$ & \small$00.00$ & \small$01.13$ \\

\cline{2-11}
& \rule{0pt}{2.3ex}\highlight{SLEM (\bm{$r$}=16)} & \highlight{\small$30.33$} & \highlight{\small69.95} & \highlight{\underline{\small$40.87$}} & \highlight{\underline{\small$47.05$}} & \highlight{\small23.12} & \highlight{\small06.60} & \highlight{\small10.40} & \highlight{\small05.49} & \highlight{\small11.40} \\

%%%%%%%%%%%%%%%%%%%%%%%%%%%%
%%% ERNIE-21B-A3B-PT
%%%%%%%%%%%%%%%%%%%%%%%%%%%%
\midrule
\midrule
\multicolumn{11}{c}{\small ERNIE-4.5-21B-A3B-PT} \\

\midrule
\multirow{2}{*}{\small 0.0\%} & Vanilla & \small$73.89$ & \small$83.30$ & \small$56.62$ & \small$71.27$ & \small$90.67$ & \small$74.80$ & \small$73.20$ & \small$89.63$ & \small$82.08$ \\
& VAE-reconstructed & \small$74.01_{\mathrlap{\pm0.15}}$ & \small$82.74_{\mathrlap{\pm0.33}}$ & \small$55.68_{\mathrlap{\pm0.46}}$ & \small$70.81_{\mathrlap{\pm0.17}}$ & \small$90.42_{\mathrlap{\pm0.59}}$ & \small$75.48_{\mathrlap{\pm1.07}}$ & \small$72.52_{\mathrlap{\pm0.41}}$ & \small$88.78_{\mathrlap{\pm1.13}}$ & \small$81.80_{\mathrlap{\pm0.35}}$ \\

\midrule
\multirow{5}{*}{\small 12.5\%} & F-Merging & \small$62.46$ & \small\textbf{82.15} & \underline{\small$52.82$} & \underline{\small$65.81$} & \small$84.69$ & \small$48.60$ & \small$36.40$ & \small$42.07$ & \small$52.94$ \\
& M-SMoE & \small$67.66$ & \small$74.73$ & \small$48.36$ & \small$63.58$ & \small$82.64$ & \small$51.60$ & \small$46.00$ & \small$51.83$ & \small$58.02$ \\
& HC-SMoE & \small\textbf{70.74} & \underline{\small$82.04$} & \small\textbf{54.61} & \small\textbf{69.13} & \small\textbf{85.90} & \underline{\small$59.00$} & \underline{\small$55.40$} & \underline{\small$62.80$} & \underline{\small$65.78$} \\
& Sub-MoE & \small$64.83$ & \small$74.46$ & \small$44.16$ & \small$61.15$ & \small$84.46$ & \small$52.80$ & \small$46.80$ & \small$60.37$ & \small$61.11$ \\

\cline{2-11}
& \rule{0pt}{2.3ex}\highlight{SLEM (\bm{$r$}=12)} & \highlight{\small$65.57$} & \highlight{\small$79.92$} & \highlight{\small$50.00$} & \highlight{\small$65.16$} & \highlight{\underline{\small$85.75$}} & \highlight{\small65.40} & \highlight{\small64.40} & \highlight{\small83.54} & \highlight{\small74.77} \\

% \textbf{
% \underline{

\midrule
\multirow{6}{*}{\small 25.0\%} & F-Merging & \underline{\small$56.92$} & \small$67.76$ & \small$40.19$ & \small$54.96$ & \small$64.82$ & \small$21.00$ & \small$11.80$ & \small$06.71$ & \small$26.08$ \\
& M-SMoE & \small$54.57$ & \small$61.45$ & \small$35.15$ & \small$50.39$ & \small$49.20$ & \small$13.20$ & \small$10.20$ & \small$07.32$ & \small$19.98$ \\
& HC-SMoE & \small\textbf{64.53} & \small\textbf{79.06} & \small\textbf{52.35} & \small\textbf{65.32} & \small\textbf{79.83} & \underline{\small$37.40$} & \underline{\small$32.20$} & \underline{\small$32.93$} & \underline{\small$45.59$} \\
& Sub-MoE & \small$51.42$ & \small$62.86$ & \small$35.34$ & \small$49.87$ & \small$50.42$ & \small$18.20$ & \small$20.20$ & \small$18.90$ & \small$26.93$ \\

\cline{2-11}
& \rule{0pt}{2.3ex}\highlight{SLEM (\bm{$r$}=12)} & \highlight{\small$53.09$} & \highlight{\underline{\small$76.26$}} & \highlight{\underline{\small$46.75$}} & \highlight{\underline{\small$58.70$}} & \highlight{\underline{\small$72.25$}} & \highlight{\small44.40} & \highlight{\small46.80} & \highlight{\small51.83} & \highlight{\small53.82} \\
\rowcolor{mainblue!10}\cellcolor{white} & $\;\llcorner$ within-layer merging & \small50.33 & \small75.42 & \small44.80 & \small56.85 & \small59.06 & \small40.00 & \small44.40 & \small45.12 & \small47.15 \\

\bottomrule
\end{tabular}}
\label{tab:main_results}
\end{table}

\section{Experiments} \label{sec:experiments}

\subsection{Experimental Setup}

\paragraph{Models and benchmarks.}
We conduct experiments on three MoE models ranging 7B-21B parameters: LFM2-8B-A1B~\citep{liquidai2025lfm2}, OLMoE-1B-7B-0125-Instruct~\citep{olmoe}, and ERNIE-4.5-21B-A3B-PT~\citep{ernie2025technicalreport}. We evaluate three categories: general knowledge reasoning with MMLU~\citep{mmlu}, ARC-Easy, and ARC-Challenge~\citep{arc}; mathematical reasoning with GSM8K~\citep{gsm8k} and MATH500~\citep{math500}; and code generation with MBPP~\citep{mbpp} and HumanEval~\citep{humaneval}. General knowledge reasoning is evaluated in an MCQA format, while the other two categories use generation-based evaluation.

\paragraph{Baselines.}
We compare SLEM against F-Merging, M-SMoE~\citep{mergecompressdemystifyefficient}, HC-SMoE~\citep{HC-SMoE}, and Sub-MoE~\citep{li2026submoe}. F-Merging uses expert activation frequencies collected from calibration data to guide the merging process. For all calibration-dependent baselines, we use randomly sampled $32\times2048$ tokens from C4 dataset~\citep{c4} to obtain the required routing statistics or expert outputs.

Due to computational constraints, we conduct experiments using greedy decoding with a single run unless otherwise specified. Further details on Expert VAE training, baseline descriptions, and benchmark configurations are provided in Appendix~\ref{app:experimental_details} and \ref{app:vae_training}.

\subsection{Functional Fidelity of Reconstructed Experts}

We first evaluate whether the trained hyperspherical VAE can effectively represent and reconstruct expert weights before applying latent-space merging. We sample latent representations from the posterior conditioned on the pretrained expert weights and decode them back into the corresponding weights. Table~\ref{tab:main_results} reports the mean and standard deviation of downstream performance over five posterior samples, where the original expert weights are replaced with VAE-reconstructed weights. Across all evaluated tasks, the reconstructed models achieve performance close to that of the original models with modest variations across samples. These results indicate that the learned latent representations retain the information necessary to recover functionally effective expert weights, supporting the use of the learned hyperspherical latent space for subsequent expert interpolation and merging.

\subsection{Comparison with Expert Merging Baselines}

Table~\ref{tab:main_results} compares SLEM with existing expert merging methods across three MoE models and two expert reduction rates. For LFM2, SLEM achieves the highest average performance on both general knowledge and generation benchmarks at 12.5\% and 25\% expert reduction. For OLMoE and ERNIE-4.5, HC-SMoE performs better on average for general knowledge benchmarks, while SLEM remains competitive. In contrast, SLEM consistently achieves the highest average generation performance across all models and expert reduction settings.

Overall, these results show that SLEM maintains comparable or better general knowledge performance while providing consistently stronger performance on generation tasks, demonstrating its effectiveness across different model architectures, scales, and compression levels.

\begin{figure}[t]
    \includegraphics[width=\linewidth]{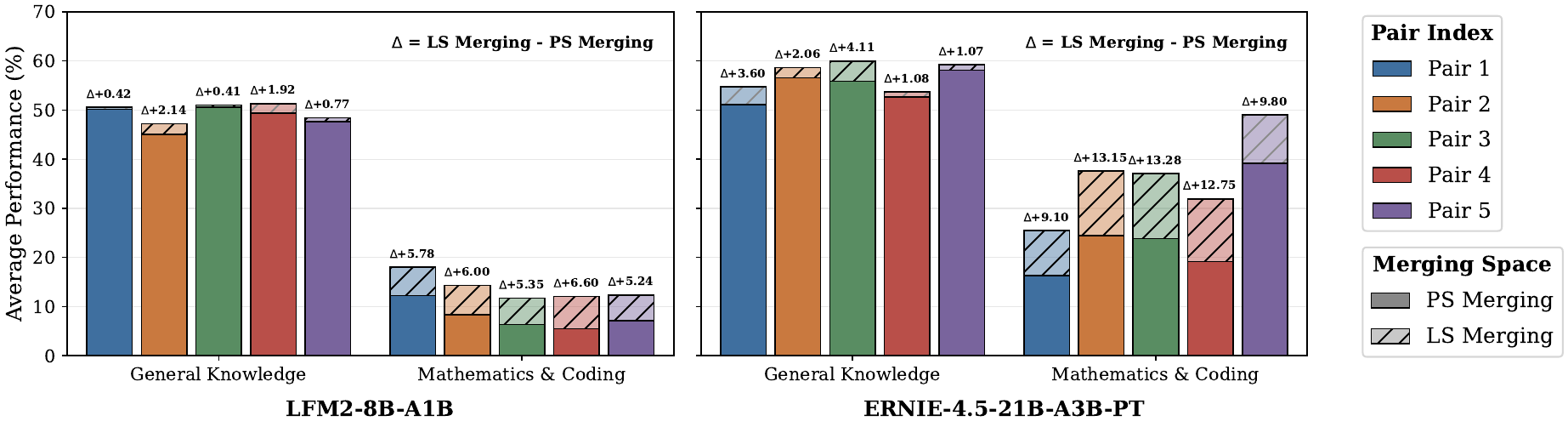}
    \caption{Comparison of parameter-space (PS) and latent-space (LS) merging under identical expert pairs and merging coefficients. Across both models and task categories, LS Merging consistently outperforms PS Merging for every evaluated pair.}
    \label{fig:merging_space}
\end{figure}
\begin{figure}[t]
    \includegraphics[width=\linewidth]{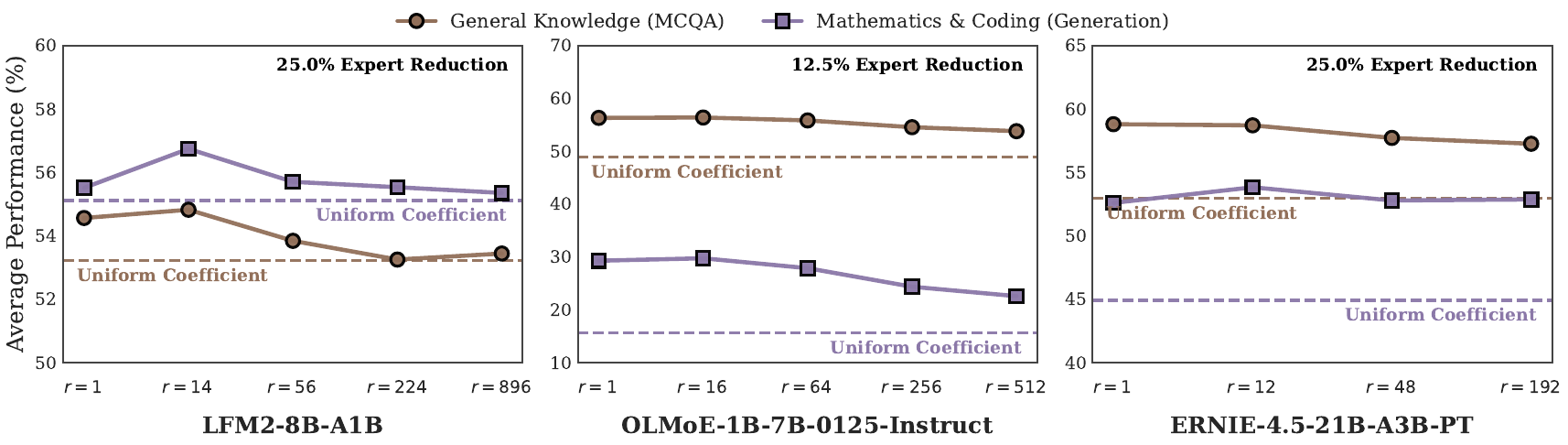}
    \caption{Effect of the truncation rank $r$ for spectrum-aware merging coefficients. Spectrum-aware coefficients consistently outperform the uniform-coefficient baseline across models and tasks.}
    \label{fig:rank_ablation}
\end{figure}

\subsection{Ablation Study}

We ablate the three key components of SLEM: merging in the learned latent space, spectrum-aware merging coefficients, and cross-layer merging.

\paragraph{Merging space.}
To isolate the effect of the merging space, we randomly construct expert pairs and assign the same uniform coefficients of $[0.5,0.5]$, varying only whether interpolation is performed in parameter space or in the learned latent space. We repeat this experiment with five independently sampled expert pairs. As shown in Figure~\ref{fig:merging_space}, latent-space merging consistently outperforms parameter-space merging for every pair across both benchmark categories and the improvement is particularly pronounced on generation-based benchmarks. These results provide empirical evidence that interpolating experts in the learned latent space offers a more effective merging space than direct interpolation of their parameters.

\paragraph{Spectrum-aware coefficients.}
We next evaluate spectrum-aware merging coefficients by varying the truncation rank $r$ used for spectral importance estimation and comparing against uniform coefficients of $[0.5,0.5]$. As shown in Figure~\ref{fig:rank_ablation}, spectrum-aware merging consistently outperforms uniform averaging across models and task categories, supporting the effectiveness of spectral importance for determining subnetwork-specific contributions. Performance also varies with truncation rank $r$, revealing a trade-off in how concentrated the resulting importance distribution becomes. As $r$ increases, the normalized SVD importance tends to become relatively uniform across subnetworks, reducing the benefit of spectrum-aware weighting. Conversely, smaller ranks produce a more discriminative importance distribution, which can improve performance. However, an overly concentrated distribution can assign negligible importance to some subnetworks, which may discard useful contributions and degrade performance. This suggests that effective merging requires a balance between distinguishing important subnetworks and preserving the contribution of overall subnetworks. We provide the normalized spectral importance distributions under different truncation ranks in Appendix~\ref{app:svd_importance_distribution}.

\begin{table}[t]
\caption{Downstream performance under constrained maximum expert group size. Limiting the group size generally degrades performance compared with unconstrained grouping.}
\centering
\resizebox{\linewidth}{!}{
\begin{tabular}{c|l c c c c c c c c c}

\toprule
\multicolumn{1}{c}{} & & \multicolumn{4}{c}{\small General Knowledge (MCQA)} & \multicolumn{5}{c}{\small Mathematics \& Coding (Generation)} \\
\cmidrule(lr){3-6} \cmidrule(lr){7-11}
\multicolumn{1}{c}{\textbf{Max Group Size}} & \textbf{Method} & \;\textbf{MMLU}\; & \;\textbf{ARC-E}\; & \;\textbf{ARC-C}\; & \textbf{Average $\uparrow$} & \textbf{GSM8K} & \textbf{MATH500} &\textbf{MBPP} & \textbf{HumanEval} & \textbf{Average $\uparrow$} \\

%%%%%%%%%%%%%%%%%%%%%%%%%%%%
%%% LFM2-8B-A1B
%%%%%%%%%%%%%%%%%%%%%%%%%%%%
\midrule
\midrule
\multicolumn{11}{c}{\small LFM2-8B-A1B} \\

\midrule
\multirow{3}{*}{$\infty$} 
& M-SMoE & \small$46.07$ & \small$66.89$ & \small$40.31$ & \small$51.09$ & \small$02.05$ & \small$02.20$ & \small$00.00$ & \small$00.00$ & \small$01.06$ \\
& HC-SMoE & \small$50.98$ & \small$60.47$ & \small$37.15$ & \small$49.53$ & \small$54.13$ & \small$45.20$ & \small$30.40$ & \small$44.51$ & \small$43.56$ \\
& Sub-MoE & \small$49.91$ & \small$64.82$ & \small$37.75$ & \small$50.83$ & \small$43.59$ & \small$34.20$ & \small$08.20$ & \small$28.05$ & \small$28.51$ \\

\midrule
\multirow{4}{*}{2} & M-SMoE & \small$34.55$ & \small$65.99$ & \small$38.99$ & \small$46.51$\down{4.58} & \small$08.34$ & \small$03.20$ & \small$00.00$ & \small$00.00$ & \small$02.89$\up{1.83} \\
& HC-SMoE & \small$49.84$ & \small$66.46$ & \small$40.10$ & \small$52.13$\up{2.60} & \small$01.52$ & \small$06.00$ & \small$01.20$ & \small$01.22$ & \small$02.49$\down{41.07} \\
& Sub-MoE & \small$47.33$ & \small$63.93$ & \small$37.12$ & \small$49.46$\down{1.37} & \small$44.20$ & \small$31.40$ & \small$08.00$ & \small$03.05$ & \small$21.66$\down{6.85} \\
\cline{2-11}
& \rule{0pt}{2.3ex}\highlight{SLEM (\bm{$r$}=14)} & \highlight{\small\textbf{51.10}} & \highlight{\small\textbf{70.62}} & \highlight{\small\textbf{42.75}} & \highlight{\small\textbf{54.82}} & \highlight{\small\textbf{70.36}} & \highlight{\small\textbf{58.80}} & \highlight{\small\textbf{36.20}} & \highlight{\small\textbf{61.59}} & \highlight{\small\textbf{56.74}} \\

%%%%%%%%%%%%%%%%%%%%%%%%%%%%
%%% ERNIE-4.5-21B-A3B-PT
%%%%%%%%%%%%%%%%%%%%%%%%%%%%
\midrule
\midrule
\multicolumn{11}{c}{\small ERNIE-4.5-21B-A3B-PT} \\

\midrule
\multirow{3}{*}{$\infty$} 
& M-SMoE & \small$54.57$ & \small$61.45$ & \small$35.15$ & \small$50.39$ & \small$49.20$ & \small$13.20$ & \small$10.20$ & \small$07.32$ & \small$19.98$ \\
& HC-SMoE & \small$64.53$ & \small$79.06$ & \small$52.35$ & \small$65.32$ & \small$79.83$ & \small$37.40$ & \small$32.20$ & \small$32.93$ & \small$45.59$ \\
& Sub-MoE & \small$51.42$ & \small$62.86$ & \small$35.34$ & \small$49.87$ & \small$50.42$ & \small$18.20$ & \small$20.20$ & \small$18.90$ & \small$26.93$ \\

\midrule
\multirow{4}{*}{2} & M-SMoE & \small$54.43$ & \small$60.48$ & \small$34.13$ & \small$49.68$\down{0.71} & \small$43.37$ & \small$11.60$ & \small$08.80$ & \small$08.54$ & \small$18.08$\down{1.90} \\
& HC-SMoE & \small\textbf{55.80} & \small$69.53$ & \small$41.30$ & \small$55.54$\down{9.78} & \small$45.26$ & \small$12.60$ & \small$09.20$ & \small$11.59$ & \small$19.66$\down{25.93} \\
& Sub-MoE & \small$47.88$ & \small$62.29$ & \small$31.82$ & \small$47.33$\down{2.54} & \small$48.82$ & \small$17.80$ & \small$20.80$ & \small$15.24$ & \small$25.67$\down{1.26} \\
\cline{2-11}
& \rule{0pt}{2.3ex}\highlight{SLEM (\bm{$r$}=12)} & \highlight{\small$53.09$} & \highlight{\small76.26} & \highlight{\small46.75} & \highlight{\small58.70} & \highlight{\small72.25} & \highlight{\small44.40} & \highlight{\small46.80} & \highlight{\small51.83} & \highlight{\small53.82} \\

\bottomrule
\end{tabular}}
\label{tab:performance_by_group_size}
\end{table}

% Effect of constraining the maximum expert group size in existing expert merging methods. Limiting the group size to two generally degrades performance compared with unconstrained grouping, with particularly large drops observed for HC-SMoE.
\begin{figure}[t]
    \includegraphics[width=\linewidth]{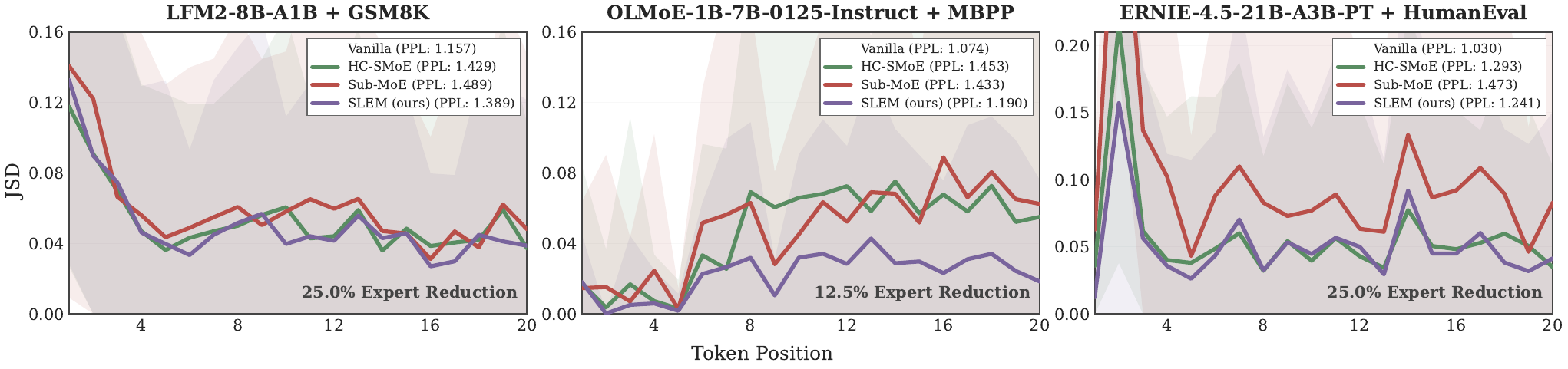}
    \caption{Token-level Jensen--Shannon divergence (JSD) and perplexity (PPL) of merged models on sequences generated by the original model. We evaluate 100 sequences only over the response tokens. Lower values indicate better preservation of the original model’s generative behavior.}
    \label{fig:token_js_divergence}
\end{figure}

\paragraph{Cross-layer merging.}
To verify the effectiveness of cross-layer merging, we compare SLEM with a within-layer variant in which the expert-grouping scope is restricted to experts from the same MoE layer, while keeping the remaining components unchanged. As shown in Table~\ref{tab:main_results}, allowing cross-layer expert pairs generally improves downstream performance across models and tasks. This suggests that the shared latent space from the trained Expert VAE provides a more flexible representation for identifying compatible experts across layer boundaries.

\subsection{Extended Analysis}

We further analyze SLEM from two perspectives. First, we examine how constraining the maximum expert group size affects existing merging methods, providing a controlled comparison under the pairwise merging setting used by SLEM. Second, we evaluate how closely the merged models preserve the generative behavior of the original model using token-level Jensen--Shannon divergence (JSD) and perplexity (PPL).

\subsubsection{Effect of Expert Group Size Constraints}

Consider a pool of $N$ experts partitioned into $M$ merge groups $\{\mathcal{G}_q\}_{q=1}^{M}$, where each group is replaced by a single merged expert. For a fixed expert reduction $R$,
\begin{equation}
R = \sum_{q=1}^{M}\bigl(|\mathcal{G}_q| - 1\bigr),
\qquad
N_{\mathrm{preserved}} = N - \sum_{q=1}^{M}|\mathcal{G}_q| = N - R - M.
\label{eq:group_size_analysis}
\end{equation}
Allowing larger groups therefore achieves the same reduction with fewer merge groups, leaving more original experts untouched. Pairwise merging, in contrast, requires $M = R$, which gives $N_{\mathrm{preserved}} = N - 2R$ and involves substantially more original experts in the merging process. This raises the question of whether effective expert merging benefits more from preserving many original experts or from limiting each merged group to few constituents. To answer it, we restrict the maximum group size of existing merging methods to two while keeping the expert reduction rate fixed.

As shown in Table~\ref{tab:performance_by_group_size}, this constraint generally degrades performance across the baselines, with particularly large drops for HC-SMoE. \citet{reap} report a high proportion of singleton clusters in HC-SMoE. We hypothesize that limiting the maximum group size reduces the number of preserved experts, which may contribute to the observed performance drop. In contrast, SLEM is designed explicitly around pairwise expert merging and maintains strong performance under the same constraint. These results suggest that SLEM can preserve useful expert functionality even when a larger fraction of the original experts must participate in merging.

\subsubsection{Preservation of Generative Behavior}

We further examine whether the strong downstream performance of SLEM corresponds to better preservation of the original model's generative behavior. Figure~\ref{fig:token_js_divergence} reports token-level JSD between the output distributions of the original and merged models, together with PPL on sequences generated by the original model. Across the evaluated model--task pairs, SLEM exhibits consistently low JSD and achieves the lowest PPL among the merged models, indicating closer agreement with the original model's token distributions. These results suggest that SLEM more faithfully preserves the generative behavior of the original model.
\section{Conclusion} \label{sec:conclusion}

In this work, we proposed \textbf{Spectrum-Aware Latent Expert Merging (SLEM)}, a calibration-free framework for MoE compression that performs expert merging in a learned latent space. SLEM represents experts across different MoE layers in a shared latent space, enabling both within- and cross-layer merging through geometry-aware interpolation. Expert pairs are selected based on latent similarity, while subnetwork-specific coefficients are derived directly from the spectrum of pretrained expert weights.

Across MoE models of different architectures and parameter scales, SLEM consistently achieves strong performance on mathematical reasoning and code generation tasks. Our ablations further demonstrate the effectiveness of latent-space merging, spectrum-aware coefficients, and cross-layer merging. In addition, SLEM more closely preserves the generative behavior of the original model. Overall, these results highlight the potential of learned expert representations and the spectral structure of expert weights for calibration-free MoE compression.
\subsection*{AI use statement} \label{sec:llm_usage}
Generative AI tools were used solely for language editing, improving the clarity and readability of the manuscript, and assisting with literature search. All research ideas, methodological design, experiments, analyses, and conclusions were developed and verified by the authors.

\subsection*{Reproducibility statement}
To support reproducibility, we detail the relevant hyperparameters throughout the Appendix. Our code and merged models will be open-sourced upon acceptance.

\subsection*{Ethics statement}
All models and datasets used in this work are open-source and publicly available. We thank those who produced, compiled, cleaned, and released them. We have cited their work appropriately and complied with all relevant licensing and usage terms. To the best of our knowledge, these models and datasets do not reveal personally identifiable information or violate copyright restrictions.

\subsection*{Acknowledgements}
This work was supported by Institute for Information \& communications Technology Planning \& Evaluation (IITP) grant funded by the Korea government (MSIT) (RS-2019-II190075, Artificial Intelligence Graduate School Program (KAIST)), National Research Foundation of Korea (NRF) grant funded by the Korea government (MSIT) (No. RS-2023-00256259), the Institute of Information \& Communications Technology Planning \& Evaluation (IITP) with a grant funded by the Ministry of Science and ICT (MSIT) of the Republic of Korea in connection with the Global AI Frontier Lab International Collaborative Research. (No. RS-2024-00469482 \& RS-2024-00509279), Institute of Information \& communications Technology Planning \& Evaluation (IITP) grant funded by the Korea government (MSIT) (No.RS-2022-II220713, Meta-learning Applicable to Real-world Problems), the “Advanced GPU Utilization Support Program” funded by the Government of the Republic of Korea (Ministry of Science and ICT), and the InnoCORE program of the Ministry of Science and ICT (No. N10250156).

\newpage
\bibliography{iclr2027_conference}
\bibliographystyle{iclr2027_conference}

\newpage
\appendix

\section{Limitations and Future Directions}
\label{app:limitations}

Compute constraints limit our evaluation to models with up to 21B parameters and fixed-budget experiments. In addition, as with other MoE compression methods, although SLEM recovers part of the performance lost through compression, the resulting compressed checkpoints do not yet fully match the performance of their uncompressed pretrained counterparts at higher expert reduction rates.

SLEM also introduces additional design and computational costs. It requires training an expert VAE before merging, which incurs additional computation and resource overhead. Moreover, the rank used for spectral importance estimation is a hyperparameter whose optimal value can vary across models and tasks. Reducing the need for model-specific rank selection and improving the efficiency of latent-space training are therefore important directions for future work.

Finally, the current expert grouping strategy is restricted to one-to-one expert pairs. This pairwise formulation limits the achievable compression under a single merging stage and does not directly support consolidating three or more experts into a single group. Extending SLEM to many-to-one merging will require grouping criteria and aggregation strategies that can characterize the compatibility and relative contributions of multiple experts jointly. Developing such group-level similarity measures and merging rules could enable higher compression ratios while preserving expert functionality.

\section{Experimental Details} \label{app:experimental_details}

We provide further details on the models, baseline implementations, and evaluation protocols used in our experiments.

\subsection{Models}

\paragraph{LFM2-8B-A1B~\citep{liquidai2025lfm2}.}
LFM2-8B-A1B contains 8.3B total parameters, with approximately 1.5B parameters activated per token. Among its 24 layers, the first two employ dense feed-forward layers and the remaining 22 use sparse MoE layers. Each MoE layer consists of 32 routed experts and activates the top-4 experts for each token. The model hidden dimension is $D=2048$, and each routed expert has an intermediate dimension of $D_{\mathrm{MoE}}=1792$.

\paragraph{OLMoE-1B-7B-0125-Instruct~\citep{olmoe}.}
OLMoE-1B-7B-0125-Instruct has 6.9B total parameters and approximately 1.3B active parameters per token. All 16 layers employ sparse MoE feed-forward networks without shared experts, each containing 64 routed experts with top-8 routing. Its hidden dimension is $D=2048$, with an expert intermediate dimension of $D_{\mathrm{MoE}}=1024$.

\paragraph{ERNIE-4.5-21B-A3B-PT~\citep{ernie2025technicalreport}.}
ERNIE-4.5-21B-A3B-PT is a 21B-parameter MoE model with approximately 3B active parameters per token. It consists of 28 layers, with the first layer using a dense feed-forward network and the remaining 27 employing MoE layers. Each MoE layer contains 64 routed experts together with two shared experts, and selects the top-6 routed experts for each token. The model uses a hidden dimension of $D=2560$ and an MoE intermediate dimension of $D_{\mathrm{MoE}}=1536$.

\subsection{Baselines}

\paragraph{M-SMoE~\citep{mergecompressdemystifyefficient}.}
M-SMoE performs routing-guided expert grouping followed by frequency-weighted parameter merging. It selects dominant experts according to their activation frequencies from the calibration set and assigns the remaining experts to the most similar dominant expert. Expert similarity is estimated from router logits collected over the calibration set:
\begin{equation}
\mathcal{S}_{\mathrm{router}}(i,j)
=
\operatorname{cos}\!\left(
\mathbf{r}_i,\mathbf{r}_j
\right),
\end{equation}
where $\mathbf{r}_i$ denotes the router-logit vector of expert $i$ over the calibration tokens. Given a resulting group $\mathcal{C}$, the hidden units of different experts are aligned through neuron permutation matching, and then experts are merged as
\begin{equation}
\widetilde{W}_{\mathcal{C},p}
=
\frac{
\sum_{i\in\mathcal{C}} f_i W_{i,p}
}{
\sum_{i\in\mathcal{C}} f_i
},
\label{eq:frequency_weighted_merging}
\end{equation}
where $f_i$ is the activation frequency of expert $i$ and $p\in\mathcal{P}$ denotes the projection type.

\paragraph{HC-SMoE~\citep{HC-SMoE}.}
HC-SMoE groups experts according to their similarity measured from expert outputs. Given a calibration set $\mathcal{D}_{\mathrm{cal}}$ containing $T$ tokens, each expert is represented by its average output:
\begin{equation}
\mathbf{o}_i
=
\frac{1}{T}
\sum_{x\in\mathcal{D}_{\mathrm{cal}}}
E_i(x).
\end{equation}
Starting from singleton expert clusters, it performs agglomerative hierarchical clustering using the distance between these representations. The distance between two clusters $\mathcal{C}_a$ and $\mathcal{C}_b$ is
\begin{equation}
d(\mathcal{C}_a,\mathcal{C}_b)
=
\frac{1}{|\mathcal{C}_a||\mathcal{C}_b|}
\sum_{i\in\mathcal{C}_a}
\sum_{j\in\mathcal{C}_b}
\|\mathbf{o}_i-\mathbf{o}_j\|_2 .
\end{equation}
The closest clusters are iteratively merged until the target number of experts is reached, and the experts within each cluster are merged according to \eqref{eq:frequency_weighted_merging}.

\paragraph{Sub-MoE~\citep{li2026submoe}.}
Sub-MoE groups functionally similar experts using their outputs on the calibration set. For a pair of experts, the functional similarity is defined as the average cosine similarity:
\begin{equation}
\mathcal{S}_{\mathrm{out}}(i,j)
=
\frac{1}{T}
\sum_{x\in\mathcal{D}_{\mathrm{cal}}}
\frac{
E_i(x)^\top E_j(x)
}{
\|E_i(x)\|_2\|E_j(x)\|_2
},
\end{equation}
which is used to form expert clusters through K-means clustering. Afterward, the experts within each cluster are aligned in a shared subspace. For projection $p$, the weight matrices of experts in a cluster $\mathcal{C}$ are jointly decomposed as
\begin{equation}
\operatorname{SVD}
\left(
[W_{i_1,p},\ldots,W_{i_{|\mathcal{C}|},p}]
\right)
=
U_p\Sigma_p
[V_{i_1,p};\ldots;V_{i_{|\mathcal{C}|},p}]^\top .
\end{equation}
The expert-specific $V$ components are then combined according to their activation frequencies,
\begin{equation}
\widetilde{V}_{\mathcal{C},p}
=
\frac{
\sum_{i\in\mathcal{C}} f_i V_{i,p}
}{
\sum_{i\in\mathcal{C}}f_i
},
\end{equation}
and the merged projection is reconstructed in the shared subspace as
\begin{equation}
\widetilde{W}_{\mathcal{C},p}
=
U_p\Sigma_p\widetilde{V}_{\mathcal{C},p}^{\top}.
\end{equation}

\paragraph{Frequency-based Merging.}
Frequency-based Merging (F-Merging) uses expert activation frequency as both the grouping and merging signal. Given a calibration set, we compute the activation frequency of each expert. Experts are then paired in a one-to-one manner by matching the most frequently activated expert with the least frequently activated expert. The selected pair are merged according to \eqref{eq:frequency_weighted_merging}, and are removed from further pairing. This procedure is repeated until the target number of experts is reached.

We do not include PuzzleMoE~\citep{zhao2026puzzlemoe} because its deployment pipeline differs substantially from the standard expert setting considered in our experiments. We also exclude NAMEx~\citep{ICLR2026_ef9f16cd} because its original formulation does not target reducing the number of stored experts under a fixed expert-reduction budget, making it not directly comparable to our compression setting.

\begin{table}[t]
    \caption{Benchmark-specific evaluation settings, including the evaluation metric, few-shot setting, maximum sequence length, and maximum generation tokens.}
    \centering
    \resizebox{\linewidth}{!}{\begin{tabular}{l c c c c}
    
    \toprule
    \multicolumn{1}{l}{\footnotesize \textbf{Task}} & \footnotesize \textbf{Metric} & \footnotesize \textbf{Few-shot} & \footnotesize \textbf{Maximum Sequence Length} & \footnotesize \textbf{Maximum Generation Tokens} \\

    \midrule
    \footnotesize \textbf{General Knowledge (MCQA)} \\
    \rule{0pt}{2.0ex}\footnotesize MMLU~\citep{mmlu} & \footnotesize Accuracy & \footnotesize 0-shot & \footnotesize $4{,}096$ & \footnotesize N/A \\
    \footnotesize ARC-Easy~\citep{arc} & \footnotesize Accuracy & \footnotesize 0-shot & \footnotesize $4{,}096$ & \footnotesize N/A \\
    \footnotesize ARC-Challenge~\citep{arc} & \footnotesize Accuracy & \footnotesize 0-shot & \footnotesize $4{,}096$ & \footnotesize N/A \\

    \midrule
    \footnotesize \textbf{Mathematics (Generation)} \\
    \rule{0pt}{2.0ex}\footnotesize GSM8K~\citep{gsm8k} & \footnotesize Exact Match & \footnotesize 8-shot (single-turn) & \footnotesize $4{,}096$ & \footnotesize $1{,}024$ \\
    \footnotesize MATH500~\citep{math500} & \footnotesize Exact Match & \footnotesize 4-shot (single-turn) & \footnotesize $4{,}096$ & \footnotesize $1{,}024$ \\

    \midrule
    \footnotesize \textbf{Coding (Generation)} \\
    \footnotesize MBPP~\citep{mbpp} & \footnotesize Pass@1 & \footnotesize 3-shot (single-turn) & \footnotesize $4{,}096$ & \footnotesize $1{,}024$ \\
    \footnotesize HumanEval~\citep{humaneval} & \footnotesize Pass@1 & \footnotesize 0-shot & \footnotesize $4{,}096$ & \footnotesize $1{,}024$ \\

    \bottomrule
    \end{tabular}}

    \label{tab:benchmark_setting}
\end{table}
% \begin{table}[t]
%     \caption{Training configures of hyperspherical expert VAE across the evaluated models.}
%     \centering
%     \resizebox{\linewidth}{!}{\begin{tabular}{l c c c c c c c c}
    
%     \toprule
%     \multicolumn{1}{l}{\small\textbf{Model}} & \small\textbf{\# layers (enc., dec.)} & \small\bm{$D_{\mathrm{\{emb,\,hidden,\,latent\}}}$} & \small\textbf{\# attn. heads} & \small\textbf{batch size} & \small\textbf{lr} & \small\textbf{weight decay} & \small\bm{$\kappa$} & \small\bm{$\beta$} \\

%     \midrule
%     \small LFM2-8B-A1B & \small 6 & \small $3{,}072$  & \small 24 & \small 4 & \small $5\times10^{-5}$ & \small $10^{-6}$ & \small $524{,}288$ & \small $10^{-3}$ \\
%     \small OLMoE-1B-7B-0125-Instruct & \small 6 & \small $3{,}072$  & \small 24 & \small 4 & \small $5\times10^{-5}$ & \small $10^{-6}$ & \small $524{,}288$ & \small $10^{-3}$ \\
%     \small ERNIE-4.5-21B-A3B-PT & \small 6 & \small $3{,}840$  & \small 30 & \small 4 & \small $5\times10^{-5}$ & \small $10^{-6}$ & \small $524{,}288$ & \small $10^{-3}$ \\

%     \bottomrule
%     \end{tabular}}

%     \label{tab:training_configures}
% \end{table}

\begin{table}[t]
    \caption{Architecture configurations of the hyperspherical expert VAE for each evaluated model. Training and optimization hyperparameters are provided in Appendix~\ref{app:vae_training}.}
    \centering
    \resizebox{\linewidth}{!}{\begin{tabular}{l c c c c c}
    
    \toprule
    \multicolumn{1}{l}{\small\textbf{Model}} & \small\textbf{\# layers (enc./dec.)} & \small\bm{$D_{\mathrm{\{emb,\,hidden,\,latent\}}}$} & \small\bm{$D_{\mathrm{FFN}}$} & \small\textbf{\# attn. heads} & \small\textbf{Pre-LN architecture} \\

    \midrule
    \small LFM2-8B-A1B & \small 6 & \small $3{,}072$ & \small $4\times3{,}072$ & \small 24 & \small yes \\
    \small OLMoE-1B-7B-0125-Instruct & \small 6 & \small $3{,}072$ & \small $4\times3{,}072$ & \small 24 & \small yes \\
    \small ERNIE-4.5-21B-A3B-PT & \small 6 & \small $3{,}840$ & \small $4\times3{,}840$ & \small 30 & \small yes \\

    \bottomrule
    \end{tabular}}

    \label{tab:training_configures}
\end{table}

\subsection{Evaluation Protocols}

We conduct all evaluations using \texttt{lm\_eval} framework~\citep{eval-harness}. For MCQA tasks, each question and its candidate options are provided as input, and the option with the highest probability is selected as the prediction. For generation-based tasks, the model generates responses autoregressively, after which the final answer is extracted using task-specific regular expressions. The benchmark-specific evaluation settings are summarized in Table~\ref{tab:benchmark_setting}. All experiments are conducted on a single NVIDIA RTX PRO 6000 Blackwell Max-Q GPU (96GB) using bfloat16 precision without quantization. Due to computational constraints, we use greedy decoding and report results from a single evaluation run unless otherwise specified.

\section{Hyperspherical Expert VAE Training} \label{app:vae_training}
We provide further details on the hyperspherical expert VAE training, including model-specific architecture configurations, training and optimization hyperparameters, and the computational resources used for training.

\paragraph{Training configurations.}
Table~\ref{tab:training_configures} summarizes the model-specific architecture configurations of the hyperspherical Expert VAE, including the number of encoder and decoder layers, representation dimensions, feed-forward dimensions, and attention heads. All models use a transformer-based encoder and decoder with a Pre-LN architecture. For training, we use a von Mises--Fisher (vMF) latent distribution with concentration parameter $\kappa=524{,}288$ and KL weight $\beta=10^{-3}$. Expert weights are standardized independently within each MoE layer, and layer-specific priors are initialized from the same randomly sampled direction. We train each VAE for 200 epochs with a batch size of 4, learning rate $5\times10^{-5}$, weight decay $10^{-6}$, and gradient clipping at 1.0. We use a cosine learning rate scheduler and bfloat16 precision training.

\paragraph{Computational resources.}
We train each Expert VAE on a single NVIDIA H200 GPU (140GB). The training cost depends on the model size and VAE configuration; for the largest evaluated model, ERNIE-4.5-21B-A3B-PT, training takes $\approx 18.7$ hours.

\section{Analysis of SLEM Design Variants}
We analyze alternative design choices of SLEM. Specifically, we compare variants that modify the scope of expert grouping and merging coefficient schemes.

\begin{table}[t]
\caption{Comparison of SLEM design variants for expert grouping and merging coefficients. We evaluate alternative grouping strategies, uniform and row-norm coefficients, and projection-specific spectral importance using only gate, up, or down projections. All variants are compared with the default SLEM under the same expert reduction setting.}
\centering
\resizebox{\linewidth}{!}{
\begin{tabular}{c|l c c c c c c c c c}

\toprule
\multicolumn{1}{c}{} & & \multicolumn{4}{c}{\small General Knowledge (MCQA)} & \multicolumn{5}{c}{\small Mathematics \& Coding (Generation)} \\
\cmidrule(lr){3-6} \cmidrule(lr){7-11}
\multicolumn{1}{c}{\textbf{Expert Reduction}} & \textbf{Method} &  \;\textbf{MMLU}\; & \;\textbf{ARC-E}\; & \;\textbf{ARC-C}\; & \textbf{\; Average $\uparrow$} & \textbf{GSM8K} & \textbf{MATH500} &\textbf{MBPP} & \textbf{HumanEval} & \textbf{Average $\uparrow$} \\

%%%%%%%%%%%%%%%%%%%%%%%%%%%%
%%% LFM2-8B-A1B
%%%%%%%%%%%%%%%%%%%%%%%%%%%%
\midrule
\midrule
\multicolumn{11}{c}{\small LFM2-8B-A1B} \\

\midrule
\multirow{8}{*}{\small 25.0\%} & SLEM ($r$=14) & \small51.10 & \small70.62 & \small42.75 & \small54.82 & \small70.36 & \small58.80 & \small36.20 & \small61.59 & \small56.74 \\

\rowcolor{gray!10}\cellcolor{white}& $\;\llcorner$ within-layer adaptive & \small53.12 & \small67.85 & \small41.81 & \small54.26\down{0.56} & \small68.84 & \small60.00 & \small22.00 & \small43.90 & \small48.69\down{8.05} \\
\rowcolor{gray!10}\cellcolor{white}& $\;\llcorner$ cross-layer adaptive & \small52.98 & \small69.19 & \small43.60 & \small55.26\up{0.44} & \small68.39 & \small60.40 & \small34.40 & \small55.49 & \small54.67\down{2.07} \\

& $\;\llcorner$ uniform coefficient & \small50.48 & \small68.01 & \small41.47 & \small53.32\down{1.50} & \small68.92 & \small58.80 & \small35.40 & \small57.32 & \small55.11\down{1.63} \\
& $\;\llcorner$ row-norm coefficient & \small50.71 & \small68.39 & \small41.47 & \small53.52\down{1.30} & \small67.70 & \small60.20 & \small35.40 & \small57.32 & \small55.15\down{1.59} \\

\rowcolor{gray!10}\cellcolor{white} & $\;\llcorner$ spectral importance: gate-only & \small52.14 & \small69.11 & \small41.04 & \small54.10\down{0.72} & \small69.83 & \small59.80 & \small30.20 & \small51.83 & \small52.91\down{3.83} \\
\rowcolor{gray!10}\cellcolor{white} & $\;\llcorner$ spectral importance: up-only & \small51.58 & \small69.74 & \small42.32 & \small54.55\down{0.27} & \small68.69 & \small59.80 & \small30.40 & \small54.88 & \small53.44\down{3.30} \\
\rowcolor{gray!10}\cellcolor{white} & $\;\llcorner$ spectral importance: down-only & \small51.47 & \small67.68 & \small41.55 & \small53.57\down{1.25} & \small71.49 & \small57.80 & \small36.20 & \small62.20 & \small56.92\up{0.18} \\

%%%%%%%%%%%%%%%%%%%%%%%%%%%%%%%%%%
%%% OLMoE-1B-7B-0125-Instruct
%%%%%%%%%%%%%%%%%%%%%%%%%%%%%%%%%%
\midrule
\midrule
\multicolumn{11}{c}{\small OLMoE-1B-7B-0125-Instruct} \\

\midrule
\multirow{8}{*}{\small 12.5\%} & SLEM ($r$=16) & \small47.37 & \small75.80 & \small45.90 & \small56.36 & \small59.74 & \small14.20 & \small20.00 & \small25.00 & \small29.74 \\

\rowcolor{gray!10}\cellcolor{white} & $\;\llcorner$ within-layer adaptive & \small46.25 & \small74.41 & \small44.88 & \small55.18\down{1.18} & \small49.58 & \small09.80 & \small18.00 & \small19.51 & \small24.22\down{5.52} \\
\rowcolor{gray!10}\cellcolor{white} & $\;\llcorner$ cross-layer adaptive & \small43.60 & \small75.13 & \small46.08 & \small54.94\down{1.42} & \small53.83 & \small10.20 & \small19.00 & \small19.51 & \small25.64\down{4.10} \\

& $\;\llcorner$ uniform coefficient & \small44.25 & \small63.89 & \small38.40 & \small48.84\down{7.52} & \small38.82 & \small08.20 & \small10.40 & \small04.88 & \small15.57\down{14.17} \\
& $\;\llcorner$ row-norm coefficient & \small46.02 & \small71.97 & \small42.32 & \small53.44\down{2.92} & \small50.87 & \small12.00 & \small15.80 & \small14.02 & \small23.17\down{6.57} \\

\rowcolor{gray!10}\cellcolor{white} & $\;\llcorner$ spectral importance: gate-only & \small47.02 & \small74.49 & \small44.80 & \small55.44\down{0.92} & \small59.44 & \small11.80 & \small21.00 & \small21.95 & \small28.55\down{1.19} \\
\rowcolor{gray!10}\cellcolor{white} & $\;\llcorner$ spectral importance: up-only & \small47.77 & \small75.80 & \small45.56 & \small56.38\up{0.02} & \small58.38 & \small14.00 & \small22.60 & \small21.95 & \small29.23\down{0.51} \\
\rowcolor{gray!10}\cellcolor{white} & $\;\llcorner$ spectral importance: down-only & \small48.07 & \small76.05 & \small45.22 & \small56.45\up{0.09} & \small58.15 & \small14.80 & \small18.60 & \small25.61 & \small29.29\down{0.45} \\

%%%%%%%%%%%%%%%%%%%%%%%%%%%%
%%% ERNIE-21B-A3B-PT
%%%%%%%%%%%%%%%%%%%%%%%%%%%%
\midrule
\midrule
\multicolumn{11}{c}{\small ERNIE-4.5-21B-A3B-PT} \\

\midrule
\multirow{8}{*}{\small 25.0\%} & SLEM ($r$=12) & \small$53.09$ & \small$76.26$ & \small$46.75$ & \small$58.70$ & \small$72.25$ & \small44.40 & \small46.80 & \small51.83 & \small53.82 \\

\rowcolor{gray!10}\cellcolor{white} & $\;\llcorner$ within-layer adaptive & \small53.27 & \small76.14 & \small43.26 & \small57.55\down{1.15} & \small63.08 & \small40.20 & \small42.40 & \small39.63 & \small46.33\down{7.49} \\
\rowcolor{gray!10}\cellcolor{white} & $\;\llcorner$ cross-layer adaptive & \small51.84 & \small74.92 & \small43.52 & \small56.76\down{1.94} & \small71.57 & \small42.00 & \small43.40 & \small53.66 & \small52.66\down{1.16} \\

& $\;\llcorner$ uniform coefficient & \small44.62 & \small73.23 & \small41.04 & \small52.96\down{5.74} & \small66.11 & \small40.60 & \small40.00 & \small32.93 & \small44.91\down{8.91} \\
& $\;\llcorner$ row-norm coefficient & \small51.46 & \small74.75 & \small43.86 & \small56.69\down{2.01} & \small71.27 & \small44.00 & \small47.60 & \small51.83 & \small53.68\down{0.14} \\

\rowcolor{gray!10}\cellcolor{white} & $\;\llcorner$ spectral importance: gate-only & \small52.86 & \small75.97 & \small44.54 & \small57.79\down{0.91} & \small71.72 & \small38.20 & \small45.20 & \small50.61 & \small51.43\down{2.39} \\
\rowcolor{gray!10}\cellcolor{white} & $\;\llcorner$ spectral importance: up-only & \small52.41 & \small74.37 & \small44.03 & \small56.94\down{1.76} & \small70.43 & \small45.20 & \small46.00 & \small46.95 & \small52.15\down{1.67} \\
\rowcolor{gray!10}\cellcolor{white} & $\;\llcorner$ spectral importance: down-only & \small54.93 & \small77.40 & \small45.39 & \small59.24\up{0.54} & \small71.49 & \small42.00 & \small42.20 & \small53.66 & \small52.34\down{1.48} \\

\bottomrule
\end{tabular}
}
\label{tab:slem_variants}
\end{table}
\subsection{Expert Grouping Variants}

SLEM constrains the maximum number of merged experts assigned to each layer, preventing merge pairs from being overly concentrated in a small subset of layers. Here, we further examine the role of this layer-wise constraint by removing the maximum merged-expert limit. We consider two variants: within-layer adaptive merging, which selects expert pairs only within each layer, and cross-layer adaptive merging, which allows pairs to be selected across layers.

As shown in Table~\ref{tab:slem_variants}, removing the layer-wise constraint generally degrades performance across the three models, with a particularly large drop for within-layer adaptive merging. We hypothesize that this degradation arises from an imbalanced distribution of merge operations across layers. Since expert redundancy, as reflected by latent similarity, is not necessarily uniform across layers, unconstrained similarity-based grouping can concentrate many merges in a small subset of layers, disproportionately reducing their expert capacity while leaving others largely unchanged. This effect is more pronounced for within-layer adaptive merging, where the candidate pool is restricted to each individual layer and highly similar local experts can be repeatedly selected. Cross-layer adaptive merging partially alleviates this issue by providing a broader pool of candidate experts, but still underperforms the balanced grouping used by SLEM. These results suggest that the layer-wise constraint acts as an effective regularizer, preserving balanced expert capacity across the layers while retaining the flexibility of cross-layer merging.

\subsection{Merging Coefficient Variants}

SLEM determines subnetwork-specific merging coefficients from the spectrum of pretrained expert weights without relying on calibration data. Here, we examine alternative calibration-free coefficient schemes to understand the contribution of spectral truncation and the aggregation of importance across projection types.

\paragraph{Uniform coefficient and row-norm coefficient.}
A simple calibration-free strategy is to assign equal contributions to the two matched subnetworks. We therefore consider a uniform coefficient of $[0.5,0.5]$. As another alternative, we use the squared row norm of each projection weight as the subnetwork importance, $s_{i,p,m}^{\mathrm{norm}}=\frac{\|W_{i,p}[m,:]\|_2^2}{\|W_{i,p}\|_F^2}$, followed by the same projection-wise averaging and coefficient computation used in SLEM. By the SVD identity, this corresponds exactly to the full-rank case of our spectral importance.

As shown in Table~\ref{tab:slem_variants}, both uniform and row-norm coefficients generally underperform our spectrum-aware coefficients. The uniform variant ignores differences in subnetwork importance entirely, while the row-norm variant incorporates the full weight energy without distinguishing dominant spectral directions. These results suggest that concentrating importance estimation on the leading singular directions provides a more informative signal for assigning the relative contributions of matched subnetworks.

\paragraph{Projection type selection.}
SLEM computes spectral importance independently for the \texttt{gate}, \texttt{up}, and \texttt{down} projections and averages them to obtain a unified subnetwork importance score. To examine the contribution of this aggregation, we evaluate variants that derive merging coefficients from only one projection type at a time.

Table~\ref{tab:slem_variants} shows that the effectiveness of each individual projection varies substantially across tasks. These results indicate that each projection captures complementary aspects of subnetwork importance. Averaging their importance scores therefore yields a more balanced estimate, resulting in more robust performance across task categories than relying on any single projection alone.

\begin{table}[t]
\caption{Downstream performance comparison with and without KL regularization during Expert VAE training. The VAE with {\small$\beta=0$} variant removes the KL divergence term while keeping the remaining training settings unchanged.}
\centering
\resizebox{\linewidth}{!}{
\begin{tabular}{c|l c c c c c c c c c}

\toprule
\multicolumn{1}{c}{} & & \multicolumn{4}{c}{\small General Knowledge (MCQA)} & \multicolumn{5}{c}{\small Mathematics \& Coding (Generation)} \\
\cmidrule(lr){3-6} \cmidrule(lr){7-11}
\multicolumn{1}{c}{\textbf{Expert Reduction}} & \textbf{Method} &  \;\textbf{MMLU}\; & \;\textbf{ARC-E}\; & \;\textbf{ARC-C}\; & \textbf{Average $\uparrow$} & \textbf{GSM8K} & \textbf{MATH500} &\textbf{MBPP} & \textbf{HumanEval} & \textbf{Average $\uparrow$} \\

%%%%%%%%%%%%%%%%%%%%%%%%%%%%
%%% LFM2-8B-A1B
%%%%%%%%%%%%%%%%%%%%%%%%%%%%
\midrule
\midrule
\multicolumn{11}{c}{\small LFM2-8B-A1B} \\

\midrule
\multirow{3}{*}{\small 0.0\%} & Vanilla & \small$63.11$ & \small$75.59$ & \small$47.63$ & \small$62.11$ & \small$81.73$ & \small$72.40$ & \small$45.60$ & \small$70.73$ & \small$67.62$ \\
& VAE-reconstructed & \small$62.93_{\mathrlap{\pm0.21}}$ & \small$75.61_{\mathrlap{\pm0.45}}$ & \small$47.41_{\mathrlap{\pm0.61}}$ & \small$61.98_{\mathrlap{\pm0.32}}$ & \small$82.35_{\mathrlap{\pm0.30}}$ & \small$72.20_{\mathrlap{\pm0.64}}$ & \small$45.52_{\mathrlap{\pm1.14}}$ & \small$71.83_{\mathrlap{\pm3.17}}$ & \small$67.97_{\mathrlap{\pm0.72}}$ \\
& $\;\llcorner$ VAE with $\beta$=0 & \small$62.94_{\mathrlap{\pm0.19}}$ & \small$75.14_{\mathrlap{\pm0.50}}$ & \small$47.25_{\mathrlap{\pm0.61}}$ & \small$61.78_{\mathrlap{\pm0.19}}$ & \small$82.02_{\mathrlap{\pm0.49}}$ & \small$72.68_{\mathrlap{\pm1.84}}$ & \small$45.56_{\mathrlap{\pm1.51}}$ & \small$69.39_{\mathrlap{\pm1.18}}$ & \small$67.41_{\mathrlap{\pm0.62}}$ \\

\midrule
\multirow{2}{*}{\small 12.5\%} & SLEM ($r$=14) & \small58.37 & \small74.79 & \small47.87 & \small60.34 & \small77.10 & \small67.60 & \small44.20 & \small62.80 & \small62.93 \\
& $\;\llcorner$ VAE with $\beta$=0 & \small54.32 & \small68.48 & \small42.41 & \small55.07\down{5.27} & \small78.54 & \small60.80 & \small29.00 & \small39.63 & \small51.99\down{10.94} \\

\midrule
\multirow{2}{*}{\small 25.0\%} & SLEM ($r$=14) & \small51.10 & \small70.62 & \small42.75 & \small54.82 & \small70.36 & \small58.80 & \small36.20 & \small61.59 & \small56.74 \\
& $\;\llcorner$ VAE with $\beta$=0 & \small44.15 & \small64.27 & \small38.48 & \small48.97\down{5.85} & \small68.76 & \small45.40 & \small17.60 & \small32.93 & \small41.17\down{15.57} \\

%%%%%%%%%%%%%%%%%%%%%%%%%%%%
%%% OLMoE-1B-7B-0125-Instruct
%%%%%%%%%%%%%%%%%%%%%%%%%%%%
\midrule
\midrule
\multicolumn{11}{c}{\small OLMoE-1B-7B-0125-Instruct} \\

\midrule
\multirow{3}{*}{\small 0.0\%} & Vanilla & \small$53.45$ & \small$76.48$ & \small$47.17$ & \small$59.03$ & \small$68.99$ & \small$23.00$ & \small$29.60$ & \small$35.37$ & \small$39.24$ \\
& VAE-reconstructed & \small$53.21_{\mathrlap{\pm0.10}}$ & \small$76.36_{\mathrlap{\pm0.56}}$ & \small$47.39_{\mathrlap{\pm0.75}}$ & \small$58.98_{\mathrlap{\pm0.32}}$ & \small$68.89_{\mathrlap{\pm0.89}}$ & \small$23.64_{\mathrlap{\pm1.08}}$ & \small$28.40_{\mathrlap{\pm0.89}}$ & \small$33.54_{\mathrlap{\pm1.02}}$ & \small$38.62_{\mathrlap{\pm0.42}}$ \\
& $\;\llcorner$ VAE with $\beta$=0 & \small$53.37_{\mathrlap{\pm0.16}}$ & \small$76.26_{\mathrlap{\pm0.51}}$ & \small$47.30_{\mathrlap{\pm1.14}}$ & \small$58.98_{\mathrlap{\pm0.53}}$ & \small$68.19_{\mathrlap{\pm0.62}}$ & \small$22.96_{\mathrlap{\pm0.48}}$ & \small$29.28_{\mathrlap{\pm1.13}}$ & \small$33.66_{\mathrlap{\pm0.98}}$ & \small$38.52_{\mathrlap{\pm0.42}}$ \\

\midrule
\multirow{2}{*}{\small 12.5\%} & SLEM (r=16) & \small47.37 & \small75.80 & \small45.90 & \small56.36 & \small59.74 & \small14.20 & \small20.00 & \small25.00 & \small29.74 \\
& $\;\llcorner$ VAE with $\beta$=0 & \small42.68 & \small71.46 & \small44.03 & \small52.72\down{3.64} & \small52.08 & \small13.60 & \small17.80 & \small15.24 & \small24.68\down{5.06} \\

\midrule
\multirow{2}{*}{\small 25.0\%} & SLEM (r=16) & \small30.33 & \small69.95 & \small40.87 & \small47.05 & \small23.12 & \small06.60 & \small10.40 & \small05.49 & \small11.40 \\
& $\;\llcorner$ VAE with $\beta$=0 & \small33.59 & \small62.12 & \small35.58 & \small43.76\down{3.29} & \small14.10 & \small03.60 & \small06.00 & \small01.22 & \small06.23\down{5.17} \\

\bottomrule
\end{tabular}}
\label{tab:kl_divergence}
\end{table}
\section{Effect of KL Regularization on Expert VAE Training}
We analyze how KL regularization during Expert VAE training affects subsequent expert merging. For this purpose, we train an additional VAE with the same posterior sampling procedure but set the KL weight to $\beta=0$, while keeping the other training configurations unchanged.

As shown in Table~\ref{tab:kl_divergence}, both VAEs achieve similar reconstruction performance on LFM2 and OLMoE. However, after latent-expert merging, the VAE trained without KL regularization shows substantially worse downstream performance. This indicates that (i) preserving reconstruction quality alone is not sufficient for effective latent interpolation, and (ii) KL regularization encourages the posterior representations to remain organized around the shared prior, while maintaining reconstruction performance. Consequently, KL regularization yields a latent space where interpolated representations remain more compatible with the decoder, leading to more effective expert merging.

\section{Paired Bootstrap Confidence Intervals}
\label{sec:bootstrap_ci}

Since SLEM shows particularly strong performance on generation-based benchmarks involving mathematical reasoning and code generation, we further assess the statistical reliability of these gains on LFM2-8B-A1B. We compare SLEM with HC-SMoE, the strongest competing baseline, using paired bootstrap confidence intervals. For each benchmark, we resample the evaluation instances with replacement while preserving the pairing between the two methods. We perform 200,000 bootstrap resamples and report the 95\% confidence interval (CI) of the performance difference in percentage points.

As shown in Table~\ref{tab:bootstrap_ci}, the confidence intervals are consistently positive for GSM8K, MATH500, and MBPP at both expert reduction rates. HumanEval includes zero at 12.5\% reduction but becomes clearly positive at 25.0\%. The confidence interval of the average generation performance also excludes zero at both reduction rates, supporting the statistical reliability of the overall generation gains of SLEM over HC-SMoE.

% \begin{table}[t]
% \caption{ci}
% \centering
% \resizebox{\linewidth}{!}{
% \begin{tabular}{c c c c c c}

% \toprule
% \multicolumn{1}{c}{} & \multicolumn{5}{c}{\small Mathematics \& Coding (Generation)} \\
% \cmidrule(lr){2-6}
% \multicolumn{1}{c}{\textbf{Expert Reduction}} & \textbf{GSM8K} & \textbf{MATH500} &\textbf{MBPP} & \textbf{HumanEval} & \textbf{Average} \\

% \midrule
% \small 12.5\% & \\
% \small 25.0\% & \\

% \bottomrule
% \end{tabular}}
% \label{tab:bootstrap}
% \end{table}

% \begin{table}[t]
% \caption{Paired bootstrap 95\% confidence intervals of the performance difference between \ours and HC-SMoE on LFM2-8B-A1B generation benchmarks. We use 200,000 bootstrap resamples. Positive intervals favor \ours.}
% \label{tab:bootstrap_ci}
% \centering
% \begin{tabular}{c l c c}
% \toprule
% \textbf{Expert Reduction} & \textbf{Benchmark} & \textbf{\# Samples} & \textbf{95\% CI (pp)} \\
% \midrule
% \multirow{5}{*}{12.5\%}
% & GSM8K     & 1,319 & [9.10, 14.33] \\
% & MATH500   & 500   & [7.60, 15.60] \\
% & MBPP      & 500   & [3.40, 11.00] \\
% & HumanEval & 164   & [-3.05, 12.80] \\
% & Average   & --    & [6.38, 11.31] \\
% \midrule
% \multirow{5}{*}{25.0\%}
% & GSM8K     & 1,319 & [13.19, 18.65] \\
% & MATH500   & 500   & [9.40, 17.80] \\
% & MBPP      & 500   & [1.00, 9.80] \\
% & HumanEval & 164   & [7.93, 23.78] \\
% & Average   & --    & [10.07, 15.32] \\
% \bottomrule
% \end{tabular}
% \end{table}

\begin{table}[h]
\caption{
Paired bootstrap 95\% confidence intervals (pp) of the performance difference between SLEM and HC-SMoE on LFM2-8B-A1B generation benchmarks.
We use 200,000 bootstrap resamples. Positive intervals favor SLEM.
}
\label{tab:bootstrap_ci}
\centering
\resizebox{0.8\linewidth}{!}{
\begin{tabular}{c c c c c c}
\toprule
\small \textbf{Expert Reduction}
& \small \textbf{GSM8K}
& \small \textbf{MATH500}
& \small \textbf{MBPP}
& \small \textbf{HumanEval}
& \small \textbf{Average} \\
&
{\small $n=1{,}319$}
& {\small $n=500$}
& {\small $n=500$}
& {\small $n=164$}
& \\
\midrule
\small12.5\%
& \small [9.10, 14.33]
& \small [7.60, 15.60]
& \small [3.40, 11.00]
& \small [-3.05, 12.80]
& \small [6.38, 11.31] \\
\small25.0\%
& \small [13.19, 18.65]
& \small [9.40, 17.80]
& \small [1.00, 9.80]
& \small [7.93, 23.78]
& \small [10.07, 15.32] \\
\bottomrule
\end{tabular}
}
\end{table}

\section{Cumulative EVR of Expert Weights}
\label{app:culmulative_evr}

\begin{figure}[t]
    \centering

    % ---------- Row 1 ----------
    \begin{subfigure}{0.32\linewidth}
        \centering
        \includegraphics[width=\linewidth]{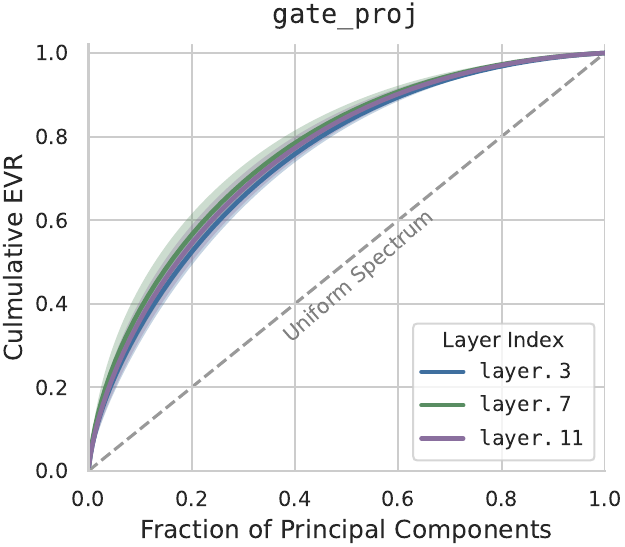}
    \end{subfigure}
    \hfill
    \begin{subfigure}{0.32\linewidth}
        \centering
        \includegraphics[width=\linewidth]{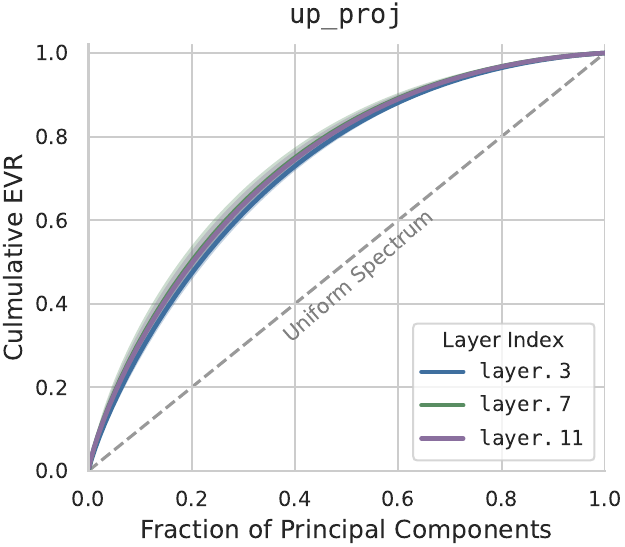}
    \end{subfigure}
    \hfill
    \begin{subfigure}{0.32\linewidth}
        \centering
        \includegraphics[width=\linewidth]{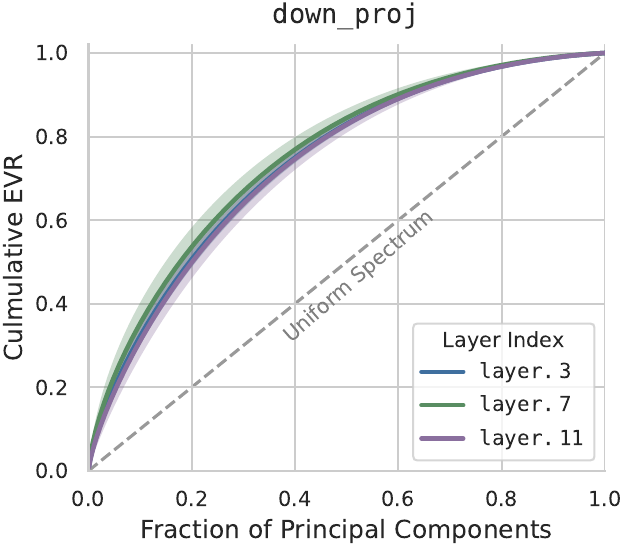}
    \end{subfigure}

    \vspace{0.3em}

    {\small (a) OLMoE-1B-7B-0125-Instruct}

    \vspace{0.7em}

    % ---------- Row 2 ----------
    \begin{subfigure}{0.32\linewidth}
        \centering
        \includegraphics[width=\linewidth]{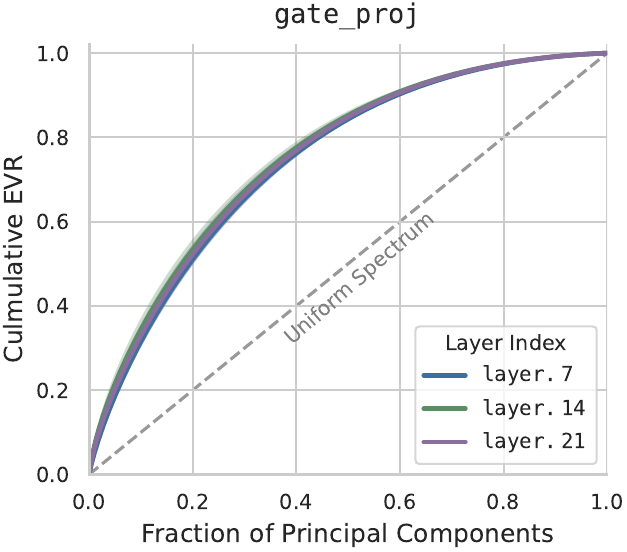}
    \end{subfigure}
    \hfill
    \begin{subfigure}{0.32\linewidth}
        \centering
        \includegraphics[width=\linewidth]{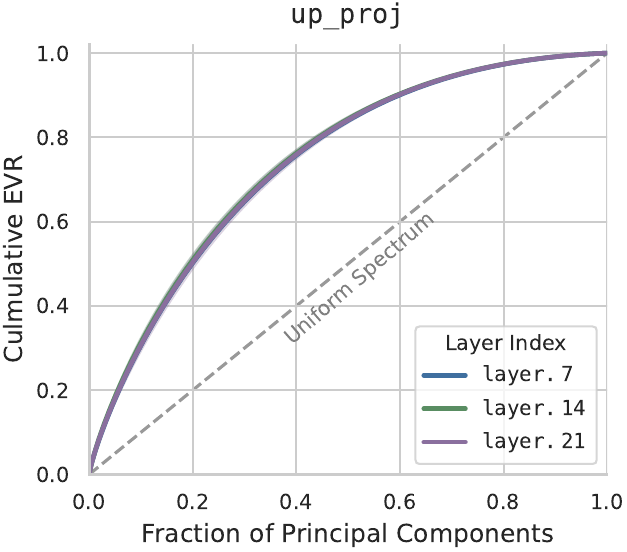}
    \end{subfigure}
    \hfill
    \begin{subfigure}{0.32\linewidth}
        \centering
        \includegraphics[width=\linewidth]{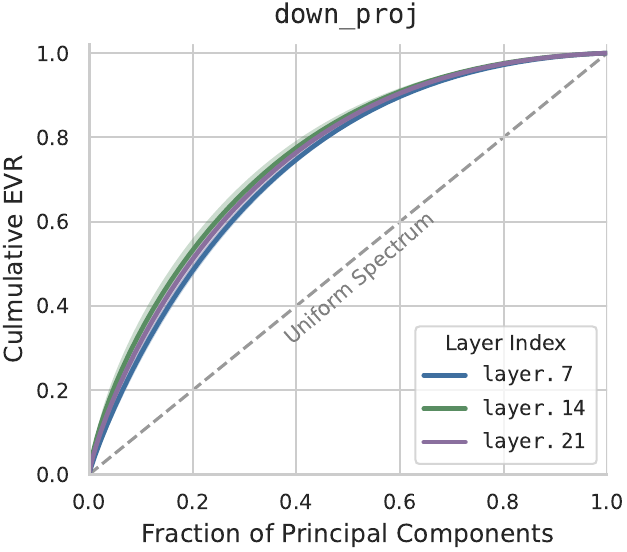}
    \end{subfigure}

    \vspace{0.3em}

    {\small (b) ERNIE-4.5-21B-A3B-PT}

    \caption{Culmulative explained variance ratio (EVR) of expert projection weights across layers in other evaluated models.}
    \label{fig:expert_evr_curve_other}
\end{figure}

We provide additional cumulative explained variance ratio (EVR) analyses for the other evaluated models. As shown in Figure~\ref{fig:expert_evr_curve_other}, their expert projection weights exhibit trends similar to those observed in the main text: variance is consistently concentrated along a subset of principal directions rather than being uniformly distributed across the spectrum. This suggests that the dominant spectral structure of expert weights is a common characteristic across the evaluated MoE models, further supporting the use of spectral information for estimating subnetwork importance.

\section{Distribution of Normalized Spectral Importance} \label{app:svd_importance_distribution}

\begin{figure}[t]
    \centering

    % ---------- Row 1 ----------
    \begin{subfigure}{\linewidth}
        \centering
        \includegraphics[width=\linewidth]{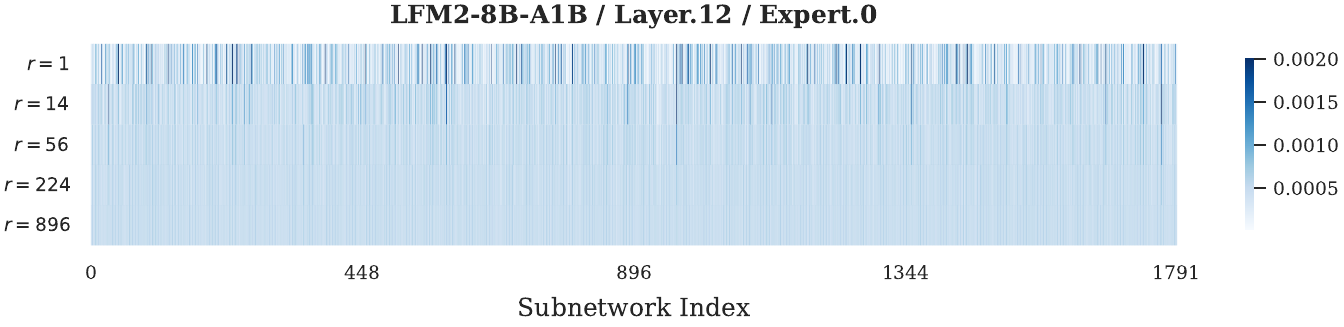}
    \end{subfigure}

    \vspace{0.5em}

    % ---------- Row 2 ----------
    \begin{subfigure}{\linewidth}
        \centering
        \includegraphics[width=\linewidth]{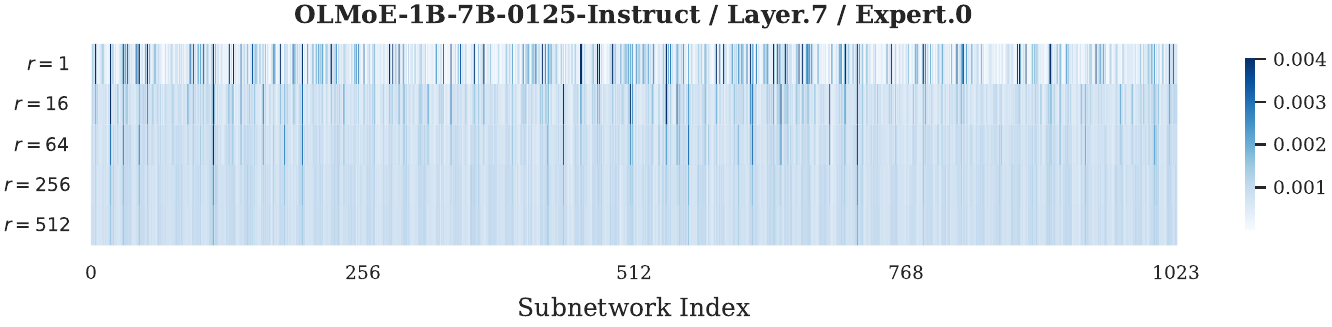}
    \end{subfigure}

    \vspace{0.5em}

    \begin{subfigure}{\linewidth}
        \centering
        \includegraphics[width=\linewidth]{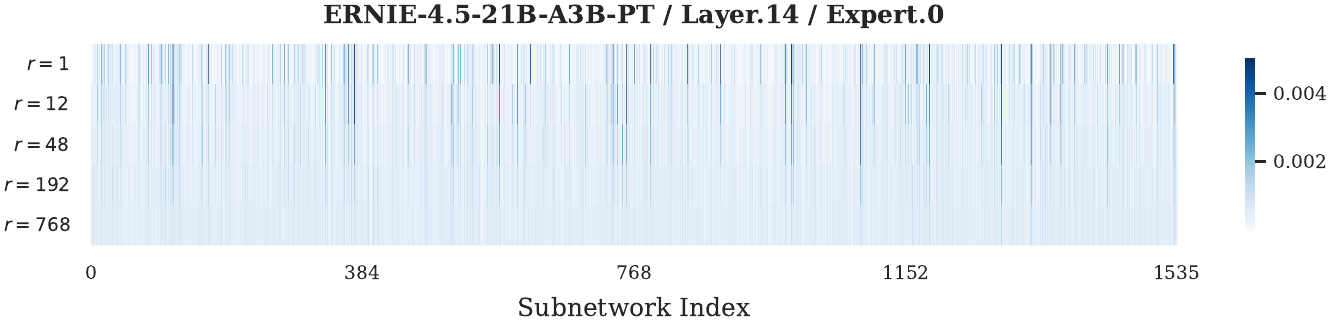}
    \end{subfigure}

    \caption{Normalized spectral importance across subnetworks under different truncation ranks $r$. Lower ranks produce more concentrated importance distributions, while larger ranks distribute importance more uniformly across subnetworks.}
    \label{fig:spectral_importance}
\end{figure}

Figure~\ref{fig:spectral_importance} visualizes the normalized spectral importance across subnetworks under different ranks $r$ for each evaluated model. As $r$ varies, the importance distribution changes in concentration, indicating that the truncation rank controls how selectively spectral energy is assigned across subnetworks.

\newpage

\end{document}